\documentclass{article}

\usepackage{times}
\usepackage{graphicx} 

\usepackage{natbib}

\usepackage{algorithm}
\usepackage{algorithmic}

\usepackage{caption}			
\usepackage{subcaption}
\usepackage{float}
\usepackage{url}

\usepackage[disable]{evolvingai} 

\usepackage[accepted]{icml2016_arxiv} 

\usepackage{enumitem}
\setlist{leftmargin=3.5mm}

\usepackage{mathtools}		  

\DeclareMathOperator*{\argmax}{arg\,max}
\newcommand{\x}{\mathbf{x}}
\newcommand{\xs}{\mathbf{x^*}}

\newcommand{\layer}[1]{\ensuremath{\mathsf{#1}\xspace}}

\usepackage[pagebackref=true,breaklinks=true,letterpaper=true,colorlinks,bookmarks=false]{hyperref}
\definecolor{mydarkblue}{rgb}{0,0.08,0.45}
\hypersetup{
	linkcolor=mydarkblue,
	citecolor=mydarkblue,
	filecolor=mydarkblue,
	urlcolor=mydarkblue,
}

\newcommand{\titl}{Multifaceted Feature Visualization: Uncovering the Different Types of Features Learned By Each Neuron in Deep Neural Networks}
\newcommand{\titlshort}{Multifaceted Feature Visualization}

\icmltitlerunning{\titlshort}

\begin{document} 

\twocolumn[
\icmltitle{\titl}

\icmlauthor{Anh Nguyen}{anguyen8@uwyo.edu}
\icmladdress{University of Wyoming}
\icmlauthor{Jason Yosinski}{yosinski@cs.cornell.edu}
\icmladdress{Cornell University}
\icmlauthor{Jeff Clune}{jeffclune@uwyo.edu}
\icmladdress{University of Wyoming}

\icmlkeywords{boring formatting information, machine learning, ICML}

\vskip 0.3in
]

\begin{abstract} 
  We can better understand deep neural networks by identifying which features each of their neurons have learned to detect. To do so, researchers have created Deep Visualization techniques including
  \emph{activation maximization}, which synthetically generates inputs (e.g. images) that maximally activate each neuron. A limitation of current techniques is that they assume each neuron detects only one type of feature, but we know that neurons can be \emph{multifaceted}, in that they fire in response to many different types of features: for example, a grocery store class neuron must activate either for rows of produce or for a storefront.
  Previous activation maximization techniques constructed images without regard for the multiple different facets of a neuron, creating inappropriate mixes of colors, parts of objects, scales, orientations, etc.
  Here, we introduce an algorithm that explicitly uncovers the multiple facets of each neuron by producing a synthetic visualization of each of the types of images that activate a neuron. We also introduce regularization methods that produce state-of-the-art results in terms of the interpretability of images obtained by activation maximization.
   By separately synthesizing each type of image a neuron fires in response to, the visualizations have more appropriate colors and coherent global structure. Multifaceted feature visualization thus provides a clearer and more comprehensive description of the role of each neuron. 
\end{abstract}

\section{Introduction}
\label{introduction}

\begin{figure}[!t]
	\centering
	\includegraphics[width=1.0\columnwidth]{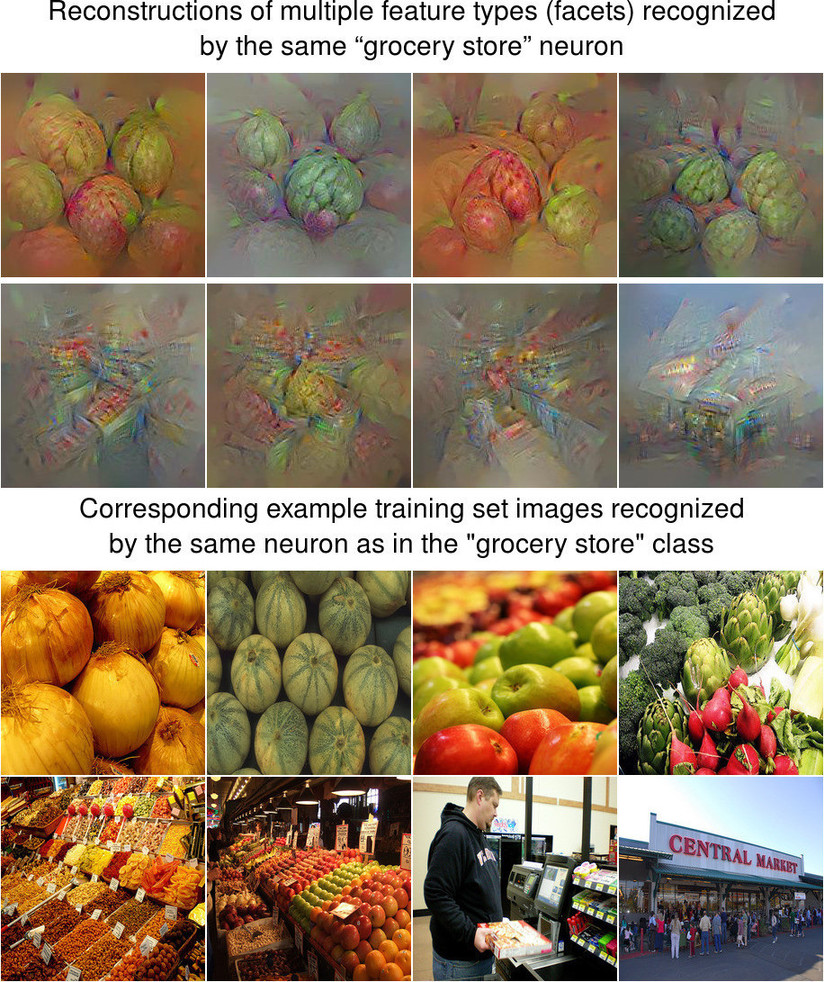}
	\caption{\textbf{Top:} Visualizations of 8 types of images (feature facets) that activate the same ``grocery store'' class neuron. \textbf{Bottom:} Example training set images that activate the same neuron, and resemble the corresponding synthetic image in the top panel.}
	\label{fig:teaser}
	\vspace{-0.5cm}
\end{figure}

Recently, deep neural networks (DNNs) have demonstrated state-of-the-art---and sometimes human-competitive---results on many pattern recognition tasks, especially vision classification problems~\cite{krizhevsky2012imagenet, szegedy2014going, karpathy2014deep,he2015deep}. That success has motivated efforts to better understand the inner workings of such networks, which enables us to further improve their architectures, learning algorithms, and hyperparameters. 
An active area of research in this vein, called \emph{Deep Visualization}, 
involves taking a trained DNN and creating synthetic images that produce specific neural activations of interest within it~\cite{zeiler2014visualizing, yosinski2015understanding,karpathy2015visualizing,dosovitskiy2015inverting,mahendran2015visualizing,nguyen2015deep,bach2015pixel,simonyan2013deep,wei2015understanding}.

There are two general camps within Deep Visualization: \emph{activation maximization}~\cite{erhan2009visualizing} and \emph{code inversion}~\cite{mahendran2015visualizing}. Activation maximization is the task of finding an image that maximally activates a certain neuron (aka ``unit'', ``feature'', or ``feature detector''), which can reveal what each neuron in a DNN has learned to fire in response to (i.e. which features it detects).
This technique can be performed for the output neurons, such as neurons that classify types of images~\cite{simonyan2013deep}, and can also be performed for each of the hidden neurons in a DNN~\cite{erhan2009visualizing,yosinski2015understanding}. 
Code inversion is the problem of synthesizing an image that, for a specific DNN layer, produces a similar activation vector at that layer as a target activation vector produced by a specific real image~\cite{mahendran2015visualizing, dosovitskiy2015inverting}. It reveals what information about one specific image is encoded by the DNN code at a particular layer.

Both activation maximization and code inversion start from a random image and calculate via backpropagation how the color of each pixel should be changed to either increase the activation of a neuron (for activation maximization) or produce a layer code closer to the target (for code inversion). However, previous studies have shown that doing so using only the gradient information produces unrealistic-looking images that are not recognizable~\cite{simonyan2013deep,nguyen2015deep}, because
the set of all possible images is so vast that it is possible to produce images that satisfy the objective, but are still unrecognizable. Instead, we must both satisfy the objective and try to limit the set of images to those that resemble natural images. Biasing optimization to produce more natural images can be accomplished by incorporating \emph{natural image priors} into optimization, which has been shown to substantially improve the recognizability of the images generated~\cite{mahendran2015visualizing,yosinski2015understanding}. Many regularization techniques have been introduced that improve image quality such as: Gaussian blur~\cite{yosinski2015understanding}, $\alpha$-norm~\cite{simonyan2013deep}, total variation~\cite{mahendran2015visualizing}, jitter~\cite{mordvintsev2015inceptionism}, and data-driven patch priors ~\cite{wei2015understanding}.

While these techniques have improved Deep Visualization methods over the last two years, the resultant images provide room for improvement: (1) the color distribution is unnatural (Fig.~\ref{fig:comparison}b-e); (2) recognizable fragments of images are repeated, but these fragments do not fit together into a coherent whole: e.g. multiple ostrich heads without bodies, or eyes without faces (Fig.~\ref{fig:comparison}b)~\cite{simonyan2013deep,mahendran2015visualizing}; (3) previous techniques have no systematic methods to visualize different facets (types of stimuli) that a neuron responds to, but high-level neurons are known to be multifaceted. For example, a face-detecting neuron in a DNN was shown to respond to both human and lion faces~\cite{yosinski2015understanding}. Neurons in human brains are similarly multifaceted: a ``Halle Berry'' neuron was found that responds to very different stimuli related to the actress, from pictures of her in costume to her name printed as text~\cite{quiroga2005invariant}.

We name the class of algorithms that visualize different facets of a neuron \emph{multifaceted feature visualization} (MFV). \citealp{erhan2009visualizing} found that optimizing an image to maximally activate a neuron from multiple random starting images usually yielded the same final visualization. In contrast, \citealp{wei2015understanding} 
found that if the backpropagation neural pathway is masked out in a certain way, the resultant images can sometimes reveal different feature facets. This is the first MFV method to our knowledge; however, it was shown to visualize only two facets per output neuron, and is not able to systematically visualize all facets per neuron. 

In this paper, we propose two Deep Visualization techniques. Most importantly, we introduce a novel MFV algorithm that:

\begin{enumerate}
	\item Sheds much more light on the inner workings of DNNs than other state-of-the-art Deep Visualization methods by revealing the different facets of each neuron. It also reveals that neurons at all levels are multifaceted, and shows that higher-level neurons are more multifaceted than lower-level 
	ones~(Fig.~\ref{fig:all_layers}).
	\item Improves the quality of synthesized images, producing state-of-the-art activation maximization results: the colors are more natural and the images are more globally consistent~(Fig.~\ref{fig:fc8_layer}. See also Figs.~\ref{fig:teaser},~\ref{fig:facets},~\ref{fig:tsne_bellpeper}) because each facet is separately synthesized. For example, MFV makes one image of a green bell pepper and another of a red bell pepper instead of trying to simultaneously make both (Fig.~\ref{fig:tsne_bellpeper}).
	The increased global structure and contextual details in the optimized images also support recent observations \cite{yosinski2015understanding} that discriminative DNNs encode not only knowledge of a sparse set of discriminative features for performing classification, but also more holistic information about typical input examples in a manner more reminiscent of generative models.
	\item Is simple to implement. The only main difference is how activation maximization algorithms are initialized. To obtain an initialization per facet, we project the training set images that maximally activate a neuron into a low-dimensional space (here, a 2D space via t-SNE), cluster the images via $k$-means, and average the $n$ (here, 15) closest images to each cluster centroid to produce the initial image (Section~\ref{sec:results}).
\end{enumerate}


We also introduce a \emph{center-biased regularization} technique that attempts to produce one central object. It combats a flaw with all activation maximization techniques including MFV: they tend to produce many repeated object fragments in an image, instead of objects with more coherent global structure. It does so by allowing, on average, more optimization iterations for center pixels than edge pixels.

\section{Methods}
\label{sec:methods}

\subsection{Convolutional neural networks}

For comparison with recent studies, we test our Deep Visualization methods on a variant of the well-known ``AlexNet'' convnet architecture \cite{krizhevsky2012imagenet}, which is trained on the 1.3-million-image ILSVRC 2012 ImageNet dataset \cite{deng2009imagenet, russakovsky2014imagenet}. Specifically, our DNN follows the CaffeNet architecture~\cite{jia2014caffe} with the weights provided by~\citealp{yosinski2015understanding}.
While this network is slightly different than AlexNet, the changes are inconsequential: the network obtains a 20.1\% top-5 error rate, which is similar to AlexNet's 18.2\%~\cite{krizhevsky2012imagenet}.

The last three layers of the DNN are fully connected: we call them \layer{fc6}, \layer{fc7} and \layer{fc8}. \layer{fc8} is the last layer (before softmax transformation) and has 1000 outputs, one for each ImageNet class. \layer{fc6} and \layer{fc7} both have 4096 outputs.

\subsection{Activation maximization}
\label{sec:act_max}

Informally, we run an optimization algorithm that optimizes the colors of the pixels in the image to maximally activate a particular neuron. That
is accomplished by calculating the derivative of the target neuron activation with respect to each pixel, which
describes how  to change the pixel color to increase the activation of that neuron. We also need to incorporate natural image priors to bias the images to remain in the set of images that look as much  as possible like natural (i.e. real-world) images. 

Formally, we may pose the activation maximization problem for a unit with index $j$ on a layer $l$ of a network $\Phi$ as finding an image $\xs$ where:

\vspace{-0.3cm}
\begin{equation}
\label{eq:activation_maximization}
\vspace{-0.3cm}
\xs = \argmax_\x(\Phi_{l,j}(\x) - R_{\theta}(\x))
\end{equation}

Here, $R_{\theta}(\x)$ is a parameterized regularization function that could include multiple regularizers (i.e. priors), each of which penalizes the search in a different way to collectively improve the image quality. We apply two regularizers: total variation (TV) and jitter, as described in~\citet{mahendran2015visualizing}, but via a slightly different, but qualitatively similar, implementation. Supplementary Sec.~\ref{sec:different_priors} details the differences and analyzes the benefits of incorporating different regularizers used in previous activation maximization papers: Gaussian blur~\cite{yosinski2015understanding}, $\alpha$-norm~\cite{simonyan2013deep}, total variation (TV)~\cite{mahendran2015visualizing}, jitter~\cite{mordvintsev2015inceptionism}, and a data-driven patch prior ~\cite{wei2015understanding}. Our code and parameters are available at \url{http://EvolvingAI.org}.

\subsection{Multifaceted feature visualization}
\label{sec:mfv}



Although DNN feature detectors must recognize that very different types of images all represent the same concept (e.g. a bell pepper detector must recognize green, red, and orange peppers as in the same class), 
there is no systematic method for visualizing these different facets of feature detectors. In this section, we introduce a method for visualizing the multiple facets that each neuron in a DNN responds to. We demonstrate the technique on the ImageNet dataset, although the results generalize beyond computer vision to any domain (e.g. speech recognition, machine translation). 

We start with the observation that in the training set, each ImageNet class has multiple intra-class clusters that reflect different facets of the class. For example, in the \emph{bell pepper} class, one finds bell peppers of different colors, alone or in groups, cut open or whole, etc. (Fig.~\ref{fig:tsne_bellpeper}). We hypothesized that activation maximization on the bell pepper class is made difficult because which facet of the bell pepper to be reconstructed is not specified, potentially meaning that different areas of the image may optimize toward reconstructing different fragments of different facets. We further hypothesized that if we initialize activation maximization optimization with the mean image of only one facet, that may increase the likelihood that optimization would reconstruct an image of that type. We also hypothesized that we can get reconstructions of different facets by starting optimization with mean images from different facets. Our experimental results support all of these hypotheses.

Specifically, to visualize different facets of an output neuron $C$ (e.g. the ~\layer{fc8} neuron that declares an image to be a member of the ``bell pepper" class), we take all (${\sim}1300$) training set examples from that class and follow Algorithm~\ref{alg:multifaceted} to produce $k=10$ facet visualizations. Note that, here, we demonstrate the algorithm on the \emph{training} set, and the results are shown in Figs.~\ref{fig:teaser},~\ref{fig:facets},~\ref{fig:tsne_bellpeper} \&~\ref{fig:fc8_layer}. However, we also apply the same technique on the \emph{validation} set (see Sec.~\ref{sec:hidden_neurons}), and show its results in Fig.~\ref{fig:all_layers}.

Intuitively, the idea is to (1) use a DNN's learned, abstract, knowledge of images to determine the different types of images that activate a neuron, and then (2) initialize a state-of-the-art activation maximization optimization algorithm with the mean image computed from each of these clusters to visualize each facet. For (1), we first embed all of the images in a class in a two-dimensional space via PCA~\cite{person1901lines,jolliffe2002principal} and t-SNE~\cite{van2008visualizing}, and then perform $k$-means clustering~\cite{macqueen1967some} to find $k$ types of images (Fig.~\ref{fig:tsne_bellpeper}). Note that here we only visualize 10 facets per neuron, but it is possible to visualize fewer or more facets by changing $k$. We compute a mean image by averaging $m=15$ images (Algorithm~\ref{alg:multifaceted}, step 5) as it works the best compared to $m=\{1, 50, 100, 200\}$ (data not shown). Supplementary Sections~\ref{sec:interpolation} \&~\ref{sec:start_mean_images} provide more intuition regarding why initializing from mean images helps.

\setlength{\textfloatsep}{0.5cm} 
\begin{algorithm}[tb]
	\caption{Multifaceted Feature Visualization}
	\label{alg:multifaceted}
	\begin{algorithmic}
		\STATE {\bfseries Input:} 
		a set of images $U$ and a number of facets $k$
		
		\small{1.}~\normalsize for each image in $U$, \textbf{compute} high-level (here \layer{fc7}) hidden code $\Phi_i$ 
		
		
		\STATE {\small{2.} \normalsize  \bfseries Reduce} the dimensionality of each code $\Phi_i$ from 4096 to 50 via PCA.

		\small{3.} \normalsize  {\bfseries Run} t-SNE visualization
		on the entire set of codes $\Phi_i$ to produce a 2-D embedding (examples in Fig.~\ref{fig:tsne_bellpeper}).
		
		\small{4.} \normalsize  {\bfseries Locate} $k$ clusters in the embedding via $k$-means.

		for each cluster
		
		~~~~\small{5.} \normalsize \textbf{Compute} a mean image $\mathbf{x_0}$ by averaging the 15
		
		~~~~images nearest to the cluster centroid.
		
		
		~~~~\small{6.} \normalsize \textbf{Run} activation maximization (see Section~\ref{sec:act_max}), but 
		
		~~~~\textbf{initialize} it with $\mathbf{x_0}$ instead of a random image.
		
		
		
		
		\STATE {\bfseries Output:} 
		a set of facet visualizations $ \{\mathbf{x_1}, \mathbf{x_2}, ..., \mathbf{x_k}\}$.
		
	\end{algorithmic}
\end{algorithm}

\subsection{Center-biased regularization}
\label{centerBiasedMainText}

In activation maximization, to increase the activation of a given neuron (e.g. an ``ostrich'' neuron), each ``drawing'' step often includes two types of updates: (1) intensifying the colors of existing ostriches in the image, and (2) drawing new fragments of ostriches (e.g. multiple heads). Since both types of updates are non-separably encoded  in the gradient, the results of previous activation maximization methods are often images with many repeated image fragments (Fig.~\ref{fig:comparison}b-h). To ameliorate this issue, we introduce a technique called \emph{center-biased regularization}.

\begin{figure}[htb]
	\centering
	\includegraphics[width=0.5\textwidth]{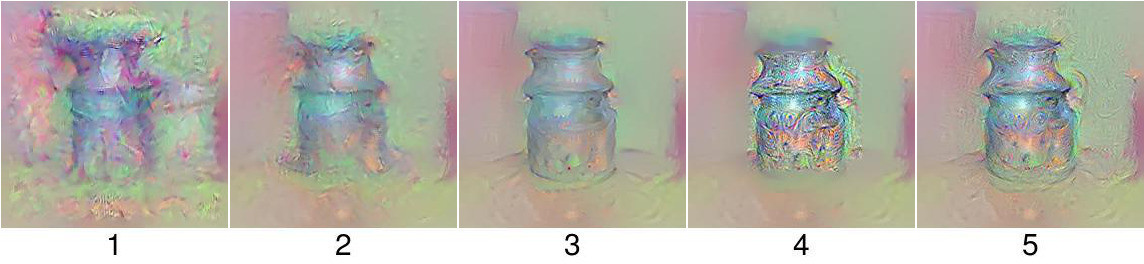}
	\vspace{-0.6cm}
	\caption{Progressive result of optimizing an image to activate the ``milk can'' neuron via the \emph{center-biased regularization} method. 
	}
	\label{fig:series}
	\vspace{-0.2cm}
\end{figure}

Preliminary experiments revealed that optimizing with a large smoothing effect (e.g. a high Gaussian blur radius or TV weight) produces a blurry, but single and centered object. Based on this observation, the intuition behind our technique is to first generate a blurry, centered object, and then refine this image by updating the center pixels more than the edge pixels to produce a final image that is sharp, and has a centrally-located object~(Fig.~\ref{fig:series}).

Center-biased regularized images have far fewer duplicated fragments (Fig.~\ref{fig:comparison}i), and thus tend to more closely represent the style of training set images, which feature one centrally located object. However, this technique is not guaranteed to produce a single object only. Instead, it is a way of biasing optimization toward creating objects near the image center. Supplementary Section~\ref{sec:center_biased} provides more details and results for center-biased regularization, including how combining it with multifaceted visualization (initializing from mean images) further improves visualization quality.

\section{Results}
\label{sec:results}

\subsection{Neurons are multifaceted feature detectors}

\begin{figure}[!t]
	\centering
	\begin{subfigure}{1.0\linewidth}
		\centering
		\includegraphics[width=1.0\linewidth]{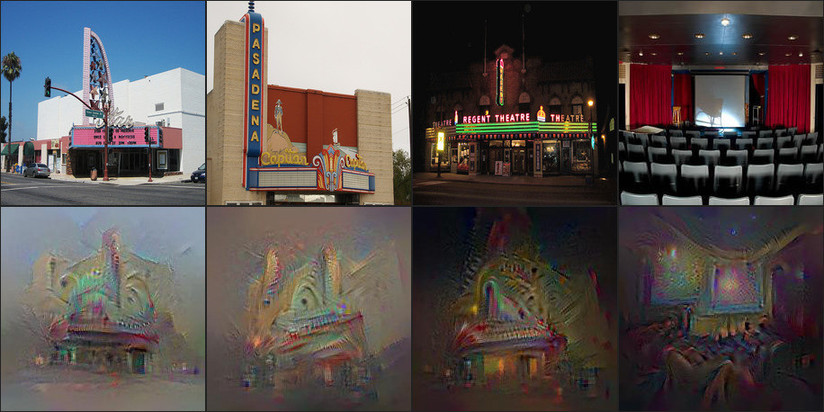}
		\caption{\emph{Movie theater}: outside (day \& night) and inside views.}
		\vspace*{1mm}
		\label{fig:facets_theater}
	\end{subfigure}
	\begin{subfigure}{1.0\linewidth}
		\centering
		\includegraphics[width=1.0\linewidth]{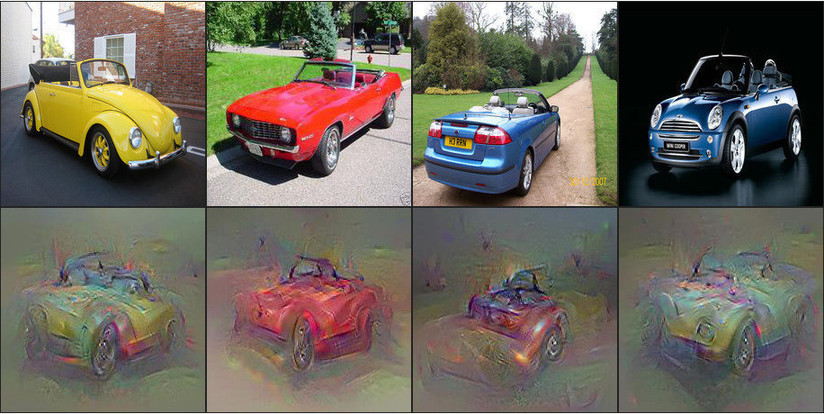}
		\caption{\emph{Convertible}: with different colors and both front \& rear views.}
		\vspace*{1mm} 		
		\label{fig:facets_convertible}
	\end{subfigure}
	\begin{subfigure}{1.0\linewidth}
		\centering
		\includegraphics[width=1.0\linewidth]{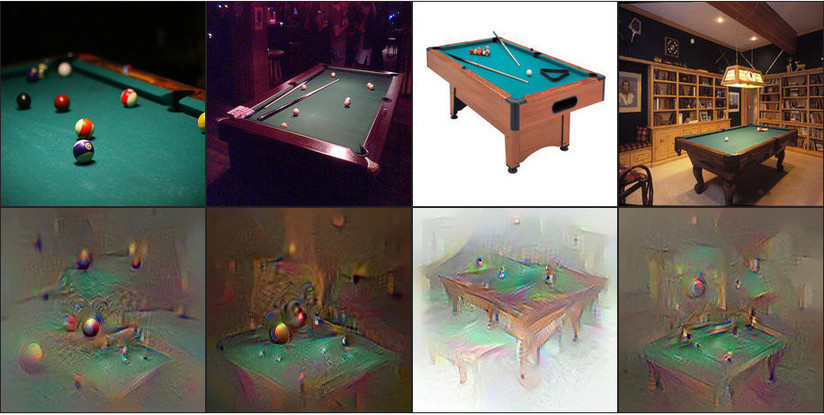}
		\caption{\emph{Pool table}: Up close \& from afar, with different backgrounds.}
		\vspace*{1mm} 		
		\label{fig:facets_pool_table}
	\end{subfigure}
	\begin{subfigure}{1.0\linewidth}
		\centering
		\includegraphics[width=1.0\linewidth]{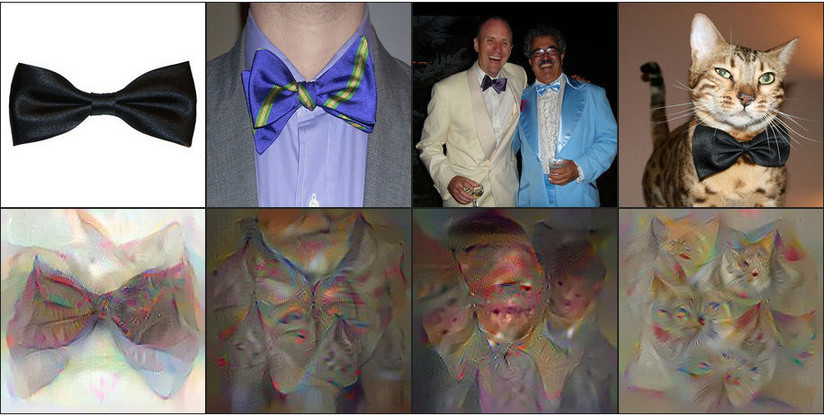}
		\caption{\emph{Bow tie}: on a white background, on one or two people, and on cats.}
		\label{fig:facets_bow_tie}
	\end{subfigure}%
	\caption{
		Multifaceted visualization of \layer{fc8} units uncovers interesting facets. We show 4 different facets for each neuron. In each pair of images, the bottom is the facet visualization that represents a cluster of images from the training set, and the top is the closest image to the visualization from the same cluster.
	}
	\label{fig:facets}
\end{figure}

\begin{figure*}
	\centering
	\includegraphics[width=2.0\columnwidth]{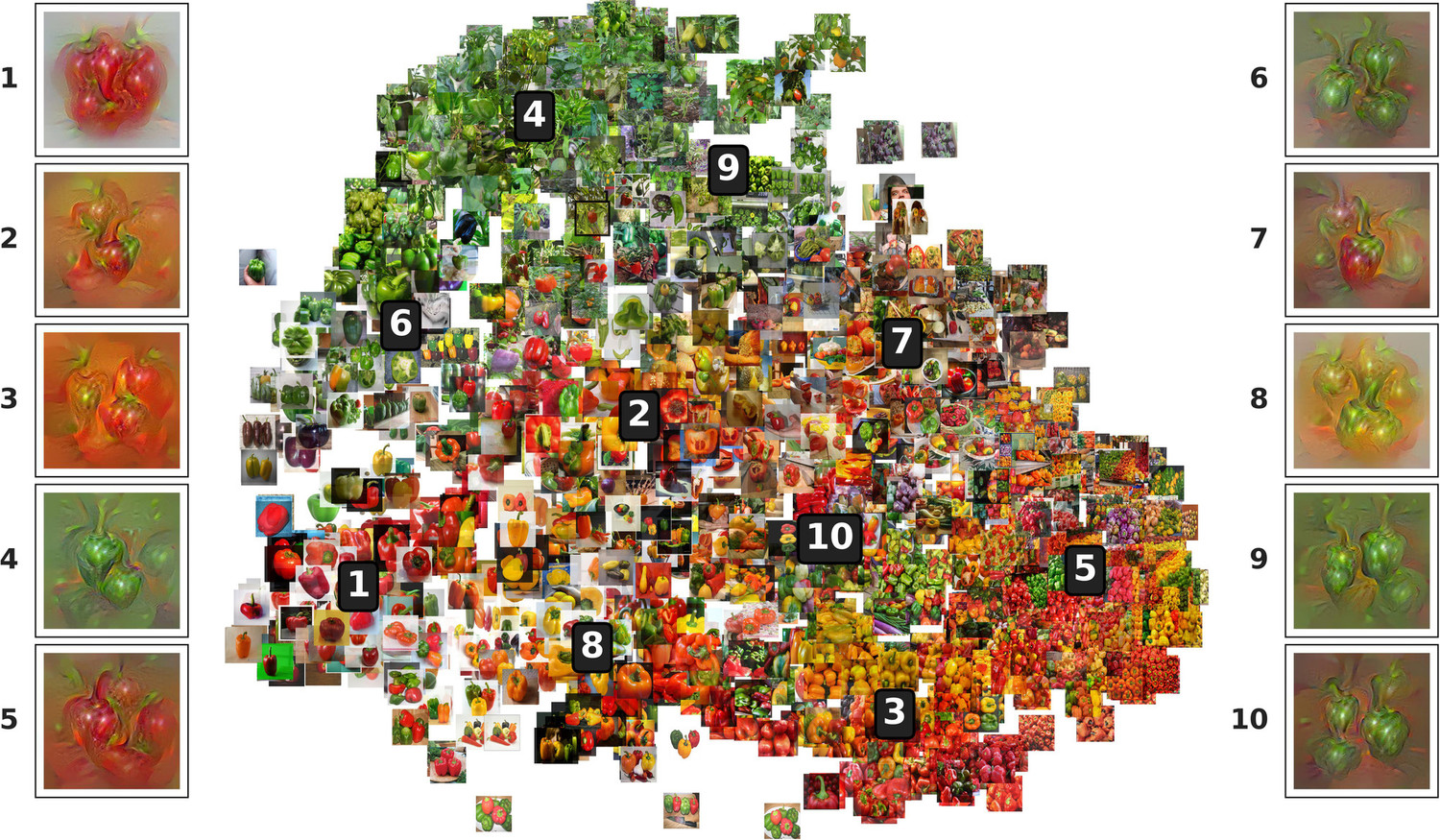}
	\vspace{-0.5em}
	\caption{
		Visualizing the different facets of a neuron that detects bell peppers. Diverse facets include a single, red bell pepper on a white background (1), multiple red peppers (5), yellow peppers (8), and green peppers on: the plant (4), a cutting board (6), or against a dark background (10). Center: training set Images from the bell pepper class are projected into two dimensions by t-SNE and clustered by $k$-means (see Sec.~\ref{sec:mfv}). Sides: synthetic images generated by multifaceted feature visualization for the ``bell pepper'' class neuron for each of the 10 numbered facets. Best viewed electronically, in color, with zoom.
		\vspace{.1cm}
	}
	\label{fig:tsne_bellpeper}
\end{figure*}

Multifaceted feature visualization produces qualitatively different images that activate the same neuron. For example, it synthesizes differently colored peppers for the ``bell pepper'' class neuron (Fig.~\ref{fig:tsne_bellpeper}) and differently colored cars for the ``convertible'' car class neuron (Fig.~\ref{fig:facets_convertible}). It also produces objects seen from different perspectives, such as cars seen from the back or front~(Fig.~\ref{fig:facets_convertible}), or in different numbers, such as one, two, or multiple peppers (Fig.~\ref{fig:tsne_bellpeper}). 
 
Most interestingly, multifaceted feature visualization can uncover that a deep neural network has learned to recognize extremely different facets of a class, such as the following. From the movie theater class: the inside of the theater with rows of seats and a stage, and the external facade viewed at night or day~(Fig.~\ref{fig:facets_theater}); from the pool table class: zoomed in pictures of pool balls on green felt, a pool table against a white background, and pool tables in dark rooms~(Fig.~\ref{fig:facets_pool_table}); from the grocery store class: rows of differently colored produce, a scene with a cashier, and the facade of the store viewed from the parking lot~(Fig.~\ref{fig:teaser});
from the bow-tie class, a tie against a white background, a single person wearing a bowtie, two people wearing bowties, and even cats wearing bowties~(Fig.~\ref{fig:facets_bow_tie}); 
in the fishing reel class, zoomed in pictures of different types of reels, pictures of reels next to fish, and pictures of people holding fishing poles against a lake or ocean background~(Fig.~\ref{fig:tsne_reel}).
See supplementary info (Fig.~\ref{fig:fc8_facets}) for many more examples.

Somewhat surprisingly, some facets of a class are, after regularized optimization, actually classified as members of other classes. For example, many real images from the training set for the grocery store class feature rows of apples or artichokes, which should arguably instead be in the apple or artichoke classes~(Fig.~\ref{fig:teaser}, bottom row). DNNs may learn to assign a high probability to both grocery store and apple for such images, and thus correctly classify the image under the ``top-5'' classification accuracy metric~\cite{krizhevsky2012imagenet}.
Multifaceted feature reconstruction correctly recognizes that different facets of the grocery store class involve each of these types of produce, and renders separate synthetic images resembling rows of apples, artichokes, or oranges. As with the real training set images, the DNN labels such synthetic images as both ``grocery store'' and the type of produce. If optimization were carried out without regularization, then following the gradient toward the grocery store class would surely assign the highest probability to that class, considering how easy it is to steer the network to produce arbitrary outputs when that is the only cost \cite{szegedy2013intriguing-properties-of-neural,nguyen2015deep,goodfellow-2014-arXiv-explaining-and-harnessing-adversarial}. However, here, when the optimization process is regularized, and initialized from the centroid of a cluster of training set images, the DNN is sometimes more confident that an image synthesized to light up its ``grocery store'' class is a member of another class (e.g. the ``apple'' class) than a member of the ``grocery store'' class!  For example, visualizations of different facets of the ``grocery store'' neuron are most confidently labeled by the DNN as in the ``custard apple'' and ``fig'' classes (Fig.~\ref{fig:teaser}, top row, leftmost two images), and only secondly labeled as in the grocery store class. These results reinforce that MFV automatically discovers and reveals the multifaceted nature of images in a class.

\subsection{Visualizing the multifaceted nature of hidden neurons}
\label{sec:hidden_neurons}

In addition to visualizing the class output neurons (layer~\layer{fc8} in this DNN), we can also perform multifaceted feature visualization for neurons on hidden layers. To do so, we collect the top $n=2\%$ (i.e. $1000$) images (or image patches, depending on the receptive field size of the neuron) from the $50,000$-image \emph{validation set} that most highly activate the neuron, which become the set of images $U$ in Algorithm~\ref{alg:multifaceted}. To determine the best $n$, we also experimented with a sweep of lower $n=\{0.05\%,0.1\%,0.15\%,0.2\%,0.25\%,0.3\%\}$, but did not observe qualitative improvements (data not shown).
To visualize a unit in a fully-connected layer $l$ (e.g. \layer{fc6}), we modify step 1 of Algorithm~\ref{alg:multifaceted} to compute the code at $l$ for each image in $U$ (e.g. a vector of 4096 numbers, one for each \layer{fc6} neuron). For a convolutional layer unit (e.g. \layer{conv3}), the layer code is the column of features (across all channels) at a given spatial location (e.g. the code for a~\layer{conv3} unit at location $(5, 7)$ is a vector of 384 numbers, one for each channel).

\begin{figure*}
	\centering
	\includegraphics[width=2.0\columnwidth]{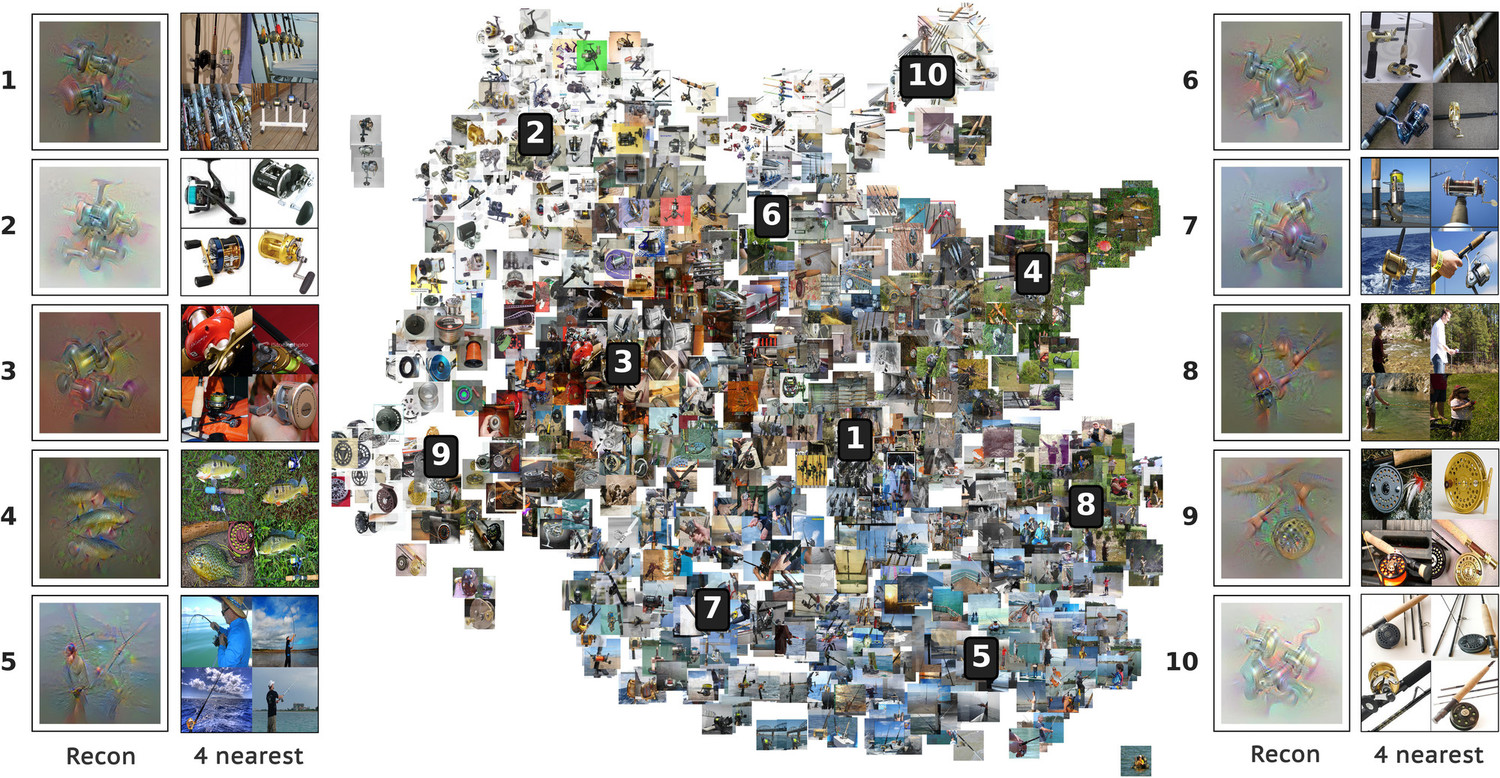}
	\vspace{-0.5em}
	\caption{
		Visualizing the different facets of a neuron that detects images in the ``fishing reel'' class. Diverse facets include reels on backgrounds that are: white (2), dark (3), ocean blue (7) or forest green (8); reels placed next to fish laying on grass (4), people fishing at sea (5), and a specific type of reel with holes in it (9).  Each reconstruction is a facet visualization for a cluster of images in the ``fishing reel'' class. The image components are as described in Fig.~\ref{fig:tsne_bellpeper}, except next to each facet visualization, we include the four images in each facet closest to the center of that facet cluster. Best viewed electronically with zoom.
	}
	\vspace{-1.5cm}
	\label{fig:tsne_reel}
\end{figure*}

\begin{figure*}
	\centering
	\includegraphics[width=1.0\textwidth]{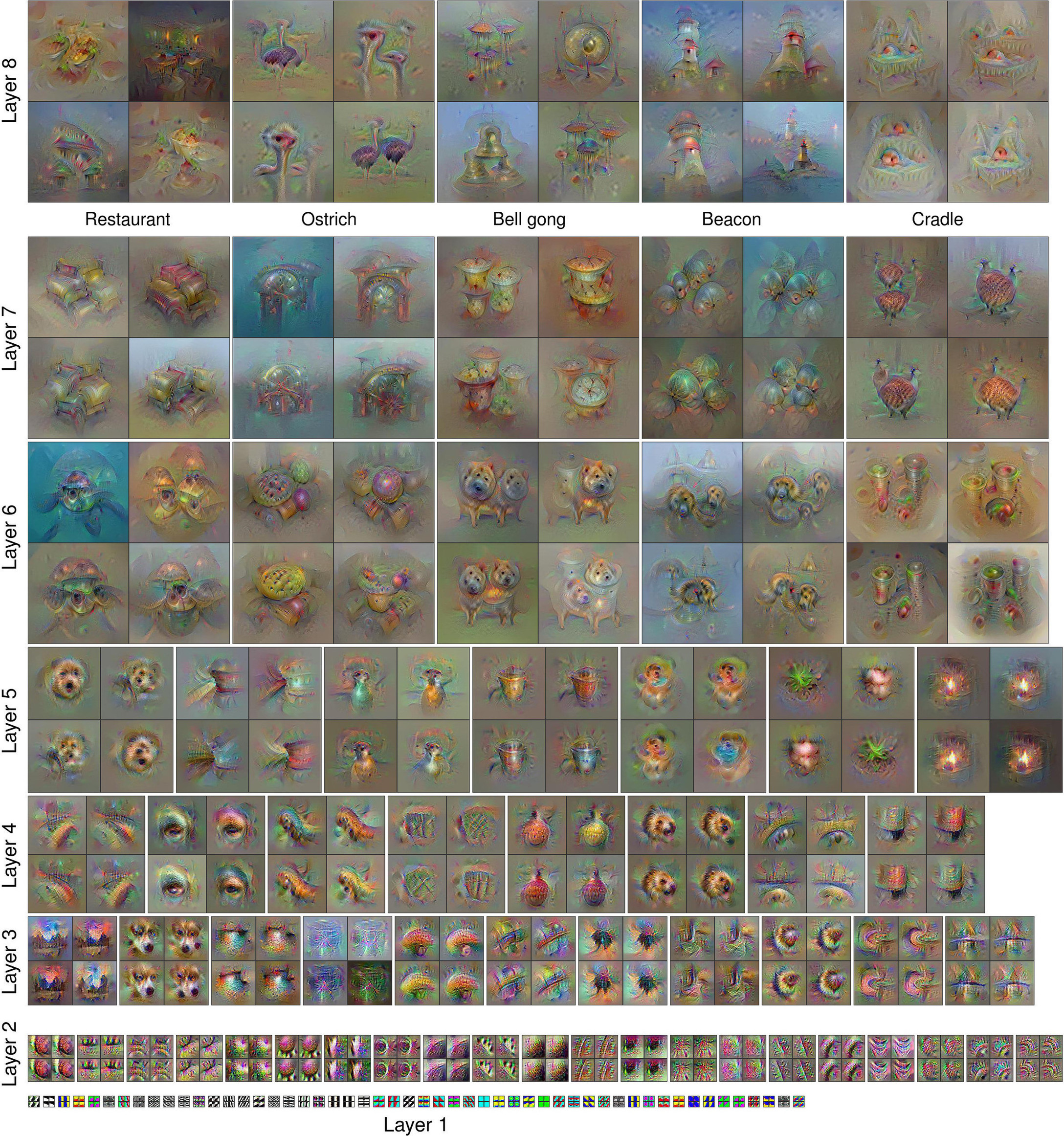}
	\caption{
		Multifaceted visualization of example neuron feature detectors from all eight layers of a deep convolutional neural network. The images reflect the true sizes of the receptive fields at different layers. For each neuron, we show visualizations of 4 different facets. These images are hand picked to showcase the diverse neurons and their multifaceted nature. Neurons in layers 1 and 2 do not have noticeably different facets. However, starting from layer 3, the visualizations reveal that neurons at higher layers are increasingly complex and have diverse facets. This increased facet diversity is because higher level features are more invariant to large changes in the image, such as color, number, pose, and context. Interestingly, units in layer 6 and 7 seem to blend different objects together. For example, the visualization of the leftmost layer 6 neuron seemingly combines a turtle and a scuba diver. It turns out this neuron responds to images of ``something underwater", including turtles and scuba divers, but also whales and sharks (Fig.~\ref{fig:fc6}). This is likely because such neurons are involved in a distributed code of abstract concepts that exist in many different classes of images. At the final, 8\textsuperscript{th}, layer, where neurons are trained to respond separately for each class, the facets are still diverse, but are within a class. Fig.~\ref{fig:fc8_facets} shows many more examples for \layer{fc8}. Best viewed electronically, in color, with zoom.
	}
	\label{fig:all_layers}
\end{figure*}

\begin{figure*}
	\centering
	\includegraphics[width=1.0\linewidth]{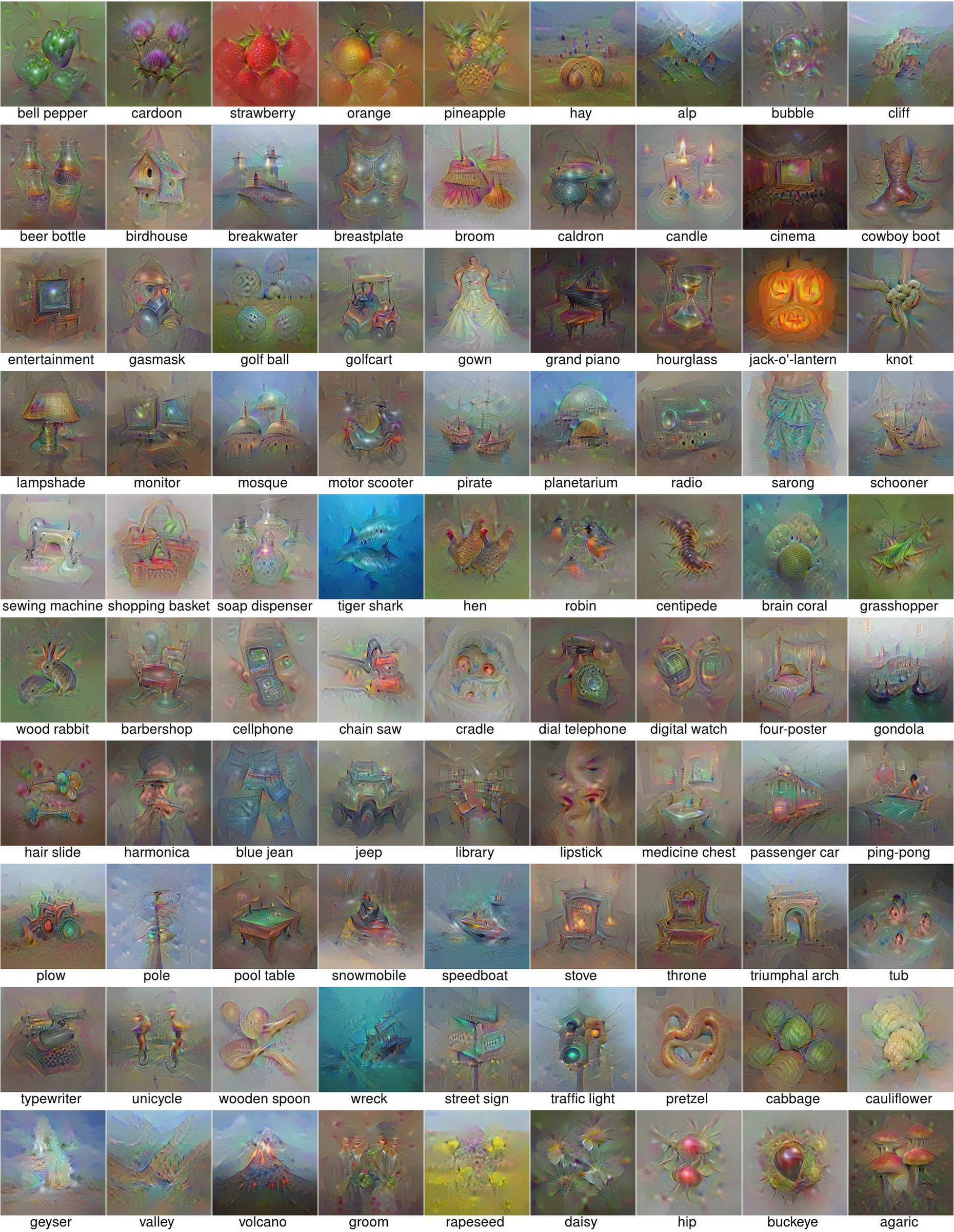}
	\caption{
		Visualizations of the facet with the most images for 90 example \layer{fc8} class neurons that showcase realistic color distributions and globally consistent objects. While subjective, we believe the improved color, detail, global consistency, and overall recognizability of these images represents the state of the art in visualization using activation maximization (Fig.~\ref{fig:comparison}).
		Moreover, the improved quality of the images reveals that even at the highest-level layer, deep neural
		networks encode much more information about classes than was previously thought, such as the global structure, details, and context of objects.
		Best viewed in color with zoom.
	}
	\label{fig:fc8_layer}
\end{figure*}

\begin{figure*}
	\centering
	\includegraphics[width=1.0\linewidth]{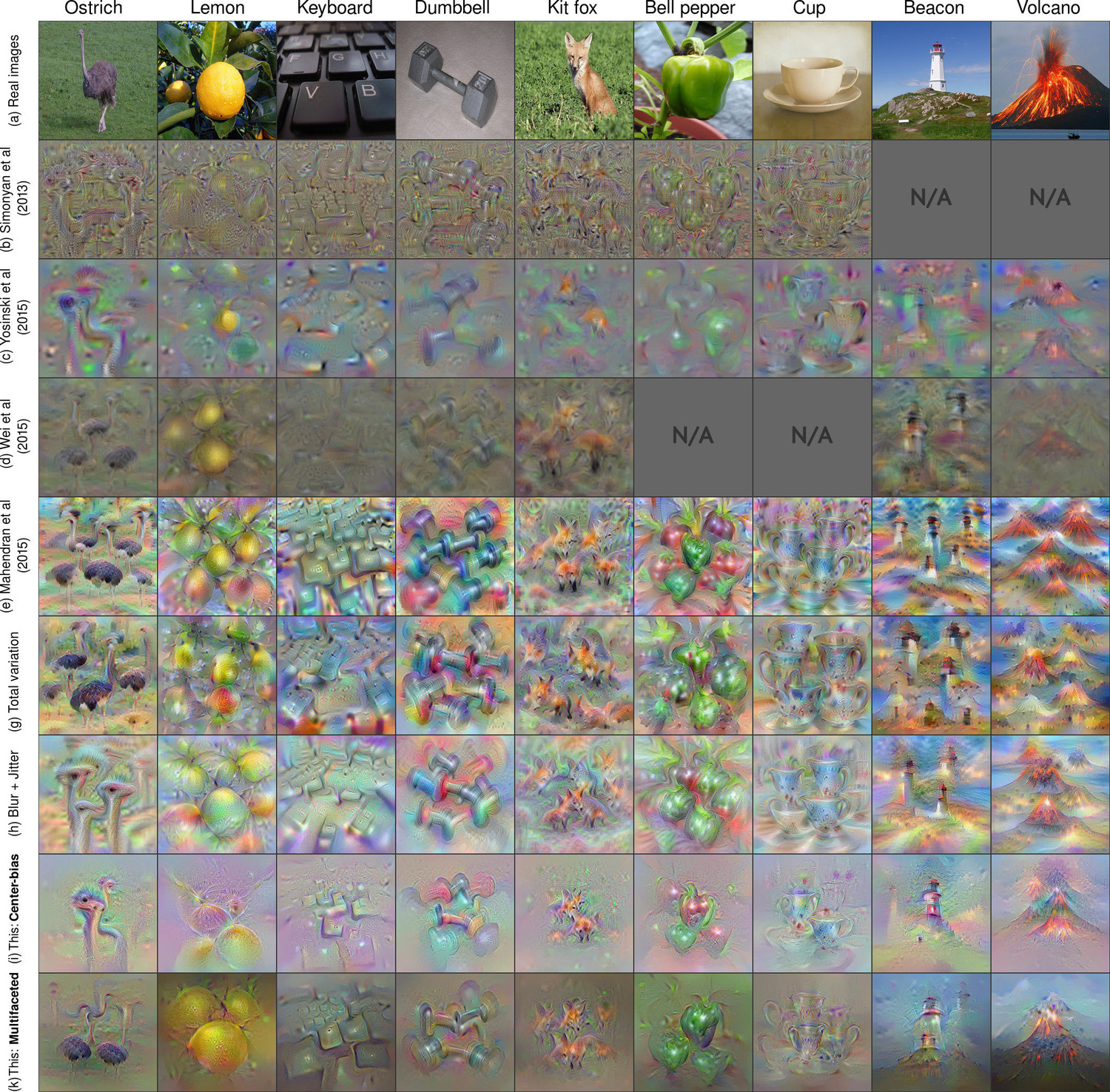}
	\caption{
		Comparing previous state-of-the-art activation maximization methods to the two new methods we propose in this paper: (k) \emph{multifaceted visualization} (Sec.~\ref{sec:mfv_improves_am}) and (i) \emph{center-biased regularization} (Sec.~\ref{centerBiasedMainText}).
		For a fair comparison, the categories were not cherry-picked to showcase our best images, but instead were selected based on the images available in previous papers~\cite{simonyan2013deep,yosinski2015understanding,wei2015understanding}. For (e), we reproduced the algorithm of~\citet{mahendran2015visualizing}, which applies both jitter and TV regularizers, although we combined them in a slightly different way (see Sec.~\ref{sec:tv}). Interestingly, we found that the visualizations produced by TV alone (g) are just as good as TV+Jitter (e); both are better than previous methods (b-d). Overall, while it is a subjective judgement and readers can decide for themselves, we believe that multifaceted feature visualization produces more recognizable images with more natural colors and more realistic global structure.
	}
	\vspace{-1.5cm}
	\label{fig:comparison}
\end{figure*}
 
Fig.~\ref{fig:all_layers} shows multiple facets reconstructed for example hidden units of every layer. Because the overall quality of these visualizations is improved (discussed more below), it becomes easier to learn what each of these hidden neurons detects. Low-level layers (e.g. \layer{conv1} and \layer{conv2}) do not exhibit noticeably different facets. However, starting at~\layer{conv3}, the qualitative difference between facets increases: first in slight differences of rotation, pose, and color, then increasingly to the number of objects present, and ultimately to very different types of objects (e.g. a closeup of an ostrich face vs. a pair of ostriches viewed from afar). This result mirrors the known phenomenon whereby features in DNNs become more abstract at higher layers, meaning that they are increasingly invariant to larger and larger changes in the input image~\cite{mahendran2015visualizing,Bengio-et-al-2015-Book}.

Another previously reported result that can be seen even more clearly in these improved visualizations is that neurons in convolutional layers often represent only a single, centered object, whereas neurons in fully connected layers are more likely to contain multiple objects~\cite{yosinski2015understanding}. 
Interestingly, and not previously reported to our knowledge, neurons in hidden, fully connected layers often seem to be an amalgam of very different, high level concepts. For example, one neuron in~\layer{fc6} looks like a combination of a scuba diver and a turtle (Fig.~\ref{fig:all_layers}, leftmost~\layer{fc6} neuron). In fact, within the top 15 images that activate that neuron, there are indeed pictures of turtles and scuba divers, but also whales and sharks. Perhaps this neuron is best described as a ``something underwater" neuron. This result could occur because our facets are not entirely pure (perhaps with a higher $k$ or different optimization objectives, we would end up with separate t-SNE clusters for underwater turtles, scuba divers, whales, and sharks). 
An alternate, but not mutually exclusive, explanation is that these feature detectors are truly amalgams, or at least represent abstract concepts (such as ``something underwater"), because they are used to classify many different types of images. It is not until the final layer that units should classify specific types of objects, and indeed our visualizations show that these last-layer (\layer{fc8}) class neurons are more pure (showing different facets of a single concept).

To further investigate these two hypotheses, we visualized the input patterns responsible for \layer{fc6} and \layer{fc7} neuron  activations via two non-stochastic methods: Deconv~\cite{zeiler2014visualizing} and Layer-wise Relevance Propagation (LRP)~\cite{bach2015pixel}. However, these methods did not reveal whether these units are truly amalgams (Sec.~\ref{sec:explain_fc_units}). Future research into these questions is necessary. 

Fully connected hidden layer neuron reconstructions also revealed cases when, given totally different initialization images (e.g. of a white photocopier and a truck), optimization still converges to the same concept (e.g. ``a yellow dog'') (Sec.~\ref{sec:explain_fc_units} \& Fig.~\ref{fig:intermed_units}). 
One might think that adding a $L_2$ penalty for deviating from the mean image in high-level (e.g. \layer{fc7}) code space might prevent such convergence and visualize more facets. Our preliminary experiments with this idea did not produce improvements, but we will continue to investigate it in future work.


\subsection{Multifaceted feature visualization improves the state of the art of activation maximization}
\label{sec:mfv_improves_am}

Beside uncovering multiple facets of a neuron, our technique also improves over the previous state-of-the-art \emph{activation maximization} methods. Figs.~\ref{fig:fc8_layer} \& \ref{fig:comparison}
showcase multifaceted feature visualization on many classes and compare it to previous methods. Note that these two figures only show one facet per class, specifically the largest among $10$ clusters (i.e. Algorithm~\ref{alg:multifaceted}, with $k=10$). While that cluster may be the largest because it contains the most canonical images, it may also be large because it is amorphous and k-means clustering could not subdivide it. Thus, our results might be even better when visualizing the most visually homogenous facet, which we will investigate in future work.


The resulting images have a substantially more natural color palette than those produced by previous algorithms (Figs.~\ref{fig:fc8_layer}~\&~\ref{fig:comparison}). That is especially evident in the background colors, which rarely change for other methods from class to class. In contrast, with multifaceted feature visualization, it is obvious when an image is set underwater or beneath a clear, blue sky. An additional improvement is that the images are more globally consistent, whereas previous methods had the problem of frequently repeated fragments of images that did not form a coherent whole. 

Initializing from mean images may work because it biases optimization towards the general color layout of a facet, providing some global structure (Fig.~\ref{fig:center_bias_only_vs_mv} shows example mean images). 
An alternate hypothesis is that the averaging operation leaves a translucent version of all $n$ original images, and optimization recovers one or several of them. Indeed, when initializing with interpolated averages between two images, optimization snaps to one or the other, instead of merging both (Fig.~\ref{fig:interpolation}). However, these reconstructions snap to the overall facet \emph{type} of the seed image--- meaning the color scheme, global structure, and theme (e.g. ostriches on a grassy plain)---but are not faithful to the details of the image (e.g. the number of ostriches). The effect is more pronounced when averaging over 15 images: optimization often ignores the dominant object outlines that remain, and instead fills in different details (e.g. at a different scale), but in a way that still fits the context of the overall color layout (e.g. Fig.~\ref{fig:center_bias_only_vs_mv}; note the cheeseburger and milk can are smaller than the mean image suggests). Such observations are not conclusive, however, and our paper motivates future research into the precise dynamics of why initializing with a mean facet image improves the quality of activation maximization. 

\section{Discussion and Conclusion}

One way to study a neuron's different feature facets is to simply perform t-SNE on the real images that maximally activate it~(Fig.~\ref{fig:tsne_bellpeper}). However, doing so does not reveal what a neuron knows about a concept or class. Based on ``fooling'' images, scientists previously concluded that DNNs trained with supervised learning ignore an object's global structure, and instead only learn a few, discriminative features per class (e.g. color or texture)~\cite{nguyen2015deep}. Our work here strengthens later findings~\cite{yosinski2015understanding,mahendran2015visualizing} showing that, in contrast, DNNs trained with supervised learning act more like generative models by learning the global structure, details, and context of objects. They also learn their multiple facets. 

Activation maximization can also reveal what a DNN ignores when classifying. A reason that our method does not always reconstruct a proper facet, e.g. of \emph{stuffed} peppers (Fig.~\ref{fig:tsne_bellpeper}, cluster 7), could be that the bell pepper neuron ignores, or lightly weights, stuffing.
The number of unique images MFV produces for a unit depends on the $k$ in $k$-means (here, manually set).
Automatically identifying the true number of facets is an important, open scientific question raised, but not answered, by this paper.

Overall, we have introduced a simple Multifaceted Feature Visualization algorithm, which (1) improves the state of the art of activation maximization by producing higher quality images with more global structure, details, context, more natural colors, and (2) shows the multiple feature facets each neuron detects, which provides a more comprehensive understanding of each neuron's function. 
We also introduced a novel \emph{center-biased regularization} technique, which reduces optimization's tendency to produce repeated object fragments and instead tends to produce one central object. 
Such improved Deep Visualization techniques will increase our understanding of deep neural networks, which will in turn improve our ability to create even more powerful deep learning algorithms.



\section*{Acknowledgements} We thank Alexey Dosovitskiy for helpful discussions, and Christopher Stanton, Joost Huizinga, Richard Yang \& Cameron Wunder for editing. Jeff Clune was supported by an NSF CAREER award (CAREER: 1453549) and a hardware donation from the NVIDIA Corporation. Jason Yosinski was supported by the NASA Space Technology Research Fellowship and NSF grant 1527232.
 
\clearpage
\bibliography{references}
\bibliographystyle{icml2016}

\clearpage

\renewcommand{\thesection}{S\arabic{section}}
\renewcommand{\thesubsection}{\thesection.\arabic{subsection}}

\newcommand{\beginsupplementary}{%
	\setcounter{table}{0}
	\renewcommand{\thetable}{S\arabic{table}}%
	\setcounter{figure}{0}
	\renewcommand{\thefigure}{S\arabic{figure}}%
	\setcounter{section}{0}
}
\newcommand{\suptitl}{Supplementary Information for:\\\titl}
\newcommand{\suptitlrunning}{Supplementary Information for: \titlshort}

\beginsupplementary


\icmltitlerunning{\suptitlrunning}

\twocolumn[
\icmltitle{\suptitl}

\icmlauthor{Anh Nguyen}{anguyen8@uwyo.edu}
\icmlauthor{Jason Yosinski}{yosinski@cs.cornell.edu}
\icmlauthor{Jeff Clune}{jeffclune@uwyo.edu}

\icmlkeywords{convolutional neural networks, neural networks, visualization}

\vskip 0.3in
]


\begin{figure*}[!tb]
	\centering
	\begin{subfigure}{1.0\textwidth}
		\centering
		\includegraphics[width=1.0\linewidth]{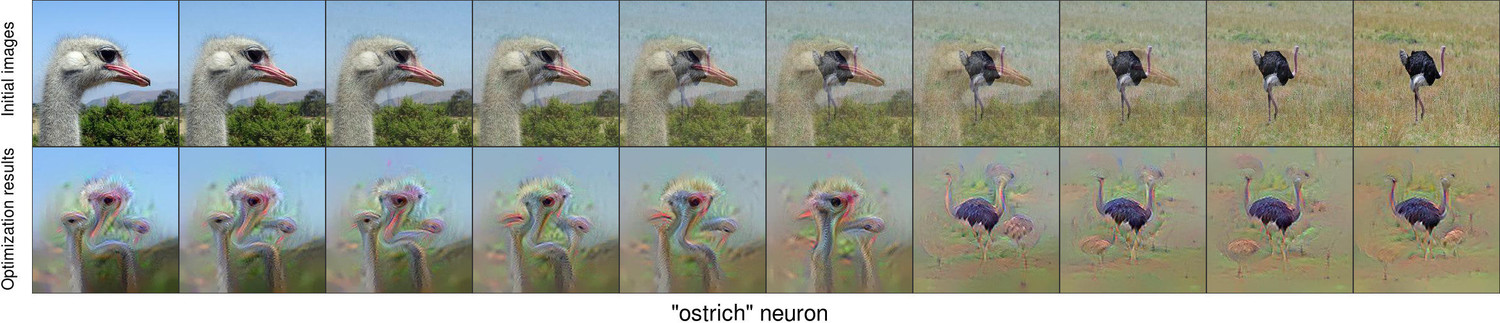}
		\vspace*{-3mm}
		\caption{Optimization is pulled toward one facet or the other instead of visualizing a combination of both. Interestingly, even when optimization is started from a single image (left and right extremes), optimization and regularization combine to produce an image in the same style (ostrich heads against a blue sky or ostriches on a grassy plain from afar), but with different details (e.g. the number of ostriches).}
		\vspace*{3mm}
		\label{fig:inter_ostrich}
	\end{subfigure}
	
	\begin{subfigure}{1.0\textwidth}
		\centering
		\includegraphics[width=1.0\linewidth]{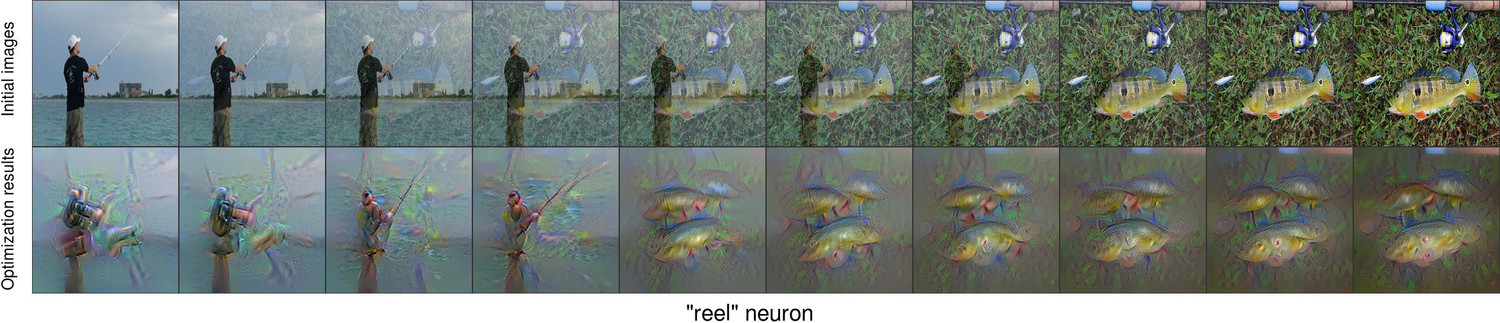}
		\vspace*{-3mm}
		\caption{Optimization either draws a fisherman against water and sky or a fish on grass. Interestingly, in the fish images, which are generated to maximize the reel class, optimization adds additional fish to the one in the seed image, instead of adding fishing reels.}
		\vspace*{3mm}
		\label{fig:inter_reel}
	\end{subfigure}
	
	\begin{subfigure}{1.0\textwidth}
		\centering
		\includegraphics[width=1.0\linewidth]{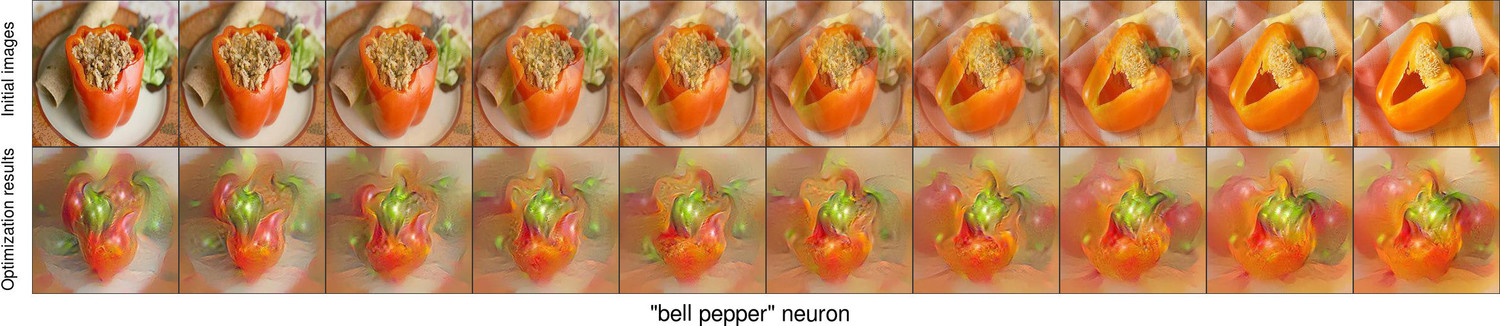}
		\vspace*{-3mm}
		\caption{Optimization seems to converge to \emph{neither} of the two facets, and instead produces a whole pepper (with inconsistent coloring). See text in Sec. \ref{sec:interpolation} for hypothesis as to why this may occur.}
		\vspace*{1.5mm}
		\label{fig:inter_bellpepper}
	\end{subfigure}	
	\caption{
		Optimization seeded with interpolated real images (top rows) often reconstructs either one facet or the other, but not a hybrid of both (bottom rows).
	}
	\label{fig:interpolation}
\end{figure*}

\begin{figure*}[tb]
	\centering
	\includegraphics[width=1.0\textwidth]{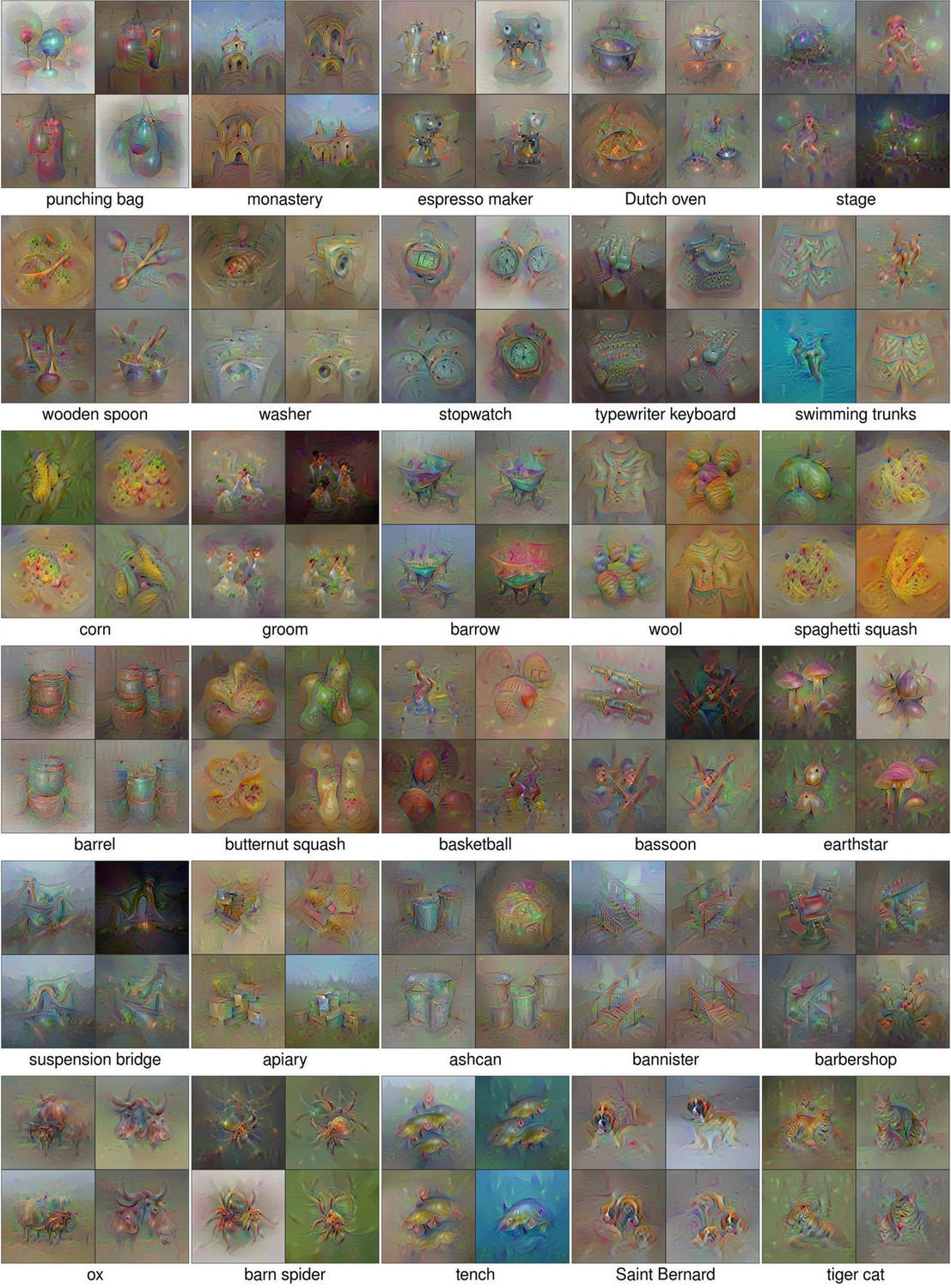}
	\caption{
		Four facets each for example class output neurons (\layer{fc8}) produced by multifaceted feature visualization. These images are hand picked to showcase the multifaceted nature of neurons across a variety of classes from natural to man-made subjects. The reconstructions are produced by applying Algorithm~\ref{alg:multifaceted} with $k=10$ on the ImageNet 2012 validation set. Best viewed electronically with zoom.
	}
	\label{fig:fc8_facets}
\end{figure*}

\section{Activation maximization initialized with interpolated images} 
\label{sec:interpolation}

We investigated whether optimization starting from a mean image that is the average of two different types of images (i.e. two different facets) would produce either (a) a visualization that resembles a mix of the two facets, or (b) a reconstruction of only one of the facets, which could occur if one facet win outs and optimization gravitates toward it, erasing the other facet. 


We first take all $\sim$1,300 of the images in a training set class (e.g. the ostrich class) and run Algorithm~\ref{alg:multifaceted} (step 1-4) with $k=10$ to produce 10 clusters of images.
We select two random images, each from a different cluster, and generate a series of 8 intermediate linearly interpolated images between those two images (e.g. in the top row of Fig.~\ref{fig:inter_ostrich}, the leftmost and rightmost are two training set images). We then run activation maximization with the same parameters as in the multifaceted visualization experiments in the main text (Sec.~\ref{sec:mfv}), with the initialization being each of the interpolated images. Each panel in Fig.~\ref{fig:interpolation} shows a series of interpolated images (top row) and the corresponding activation maximization results (bottom row).

We found that when starting from a single real image, sometimes optimization produces a visualization that fits in the same facet as the initial image, but with different details. For example, in Fig.~\ref{fig:inter_ostrich}, the leftmost and rightmost synthetic images are started with the (non-modified) training set image above them and reproduce a similar \emph{type} of image (large ostrich heads against a blue sky or ostriches on a grassy plain from afar), but the details differ. Because of regularization (TV and jitter) and optimization, activation maximization completely redraws the main subject, and actually produces more than one ostrich in both cases (zoom in to see more easily). 

In other cases, optimization does not take the guide from the initial image, but instead converges to a different facet altogether. For example, when seeded with either a stuffed or cut-open pepper, optimization instead seemingly produces a whole pepper (Fig.~\ref{fig:inter_bellpepper}).
This could happen because a ``stuffed pepper'' facet and a regular ``whole pepper'' facet share a lot of common details (reflective skin, colors), thus, the bias in the seed image might not be strong enough to pull the optimization toward the stuffed pepper facet. Another hypothesis is that the DNN has never learned to associate the stuffing detail with the bell pepper concept. A final hypothesis is that whole peppers are much more common, making their basin of attraction much larger and the gradients toward them steeper. 

Overall, we observe that optimization often reconstructs one facet or another, but not a hybrid of both (Figs.~\ref{fig:inter_ostrich} \&~\ref{fig:inter_reel}). Interestingly, and for unknown reasons, in the fishing reel example, optimization did not produce the `fisherman against water and sky from afar' facet until the 3rd interpolated image (Fig.~\ref{fig:inter_reel}, 3rd image from the left). Instead, in the leftmost two images of Fig.~\ref{fig:inter_reel}, optimization produced a totally different, and common, facet of a close-up of a fishing reel (Fig.~\ref{fig:tsne_reel}, most facets).

\section{Comparison between different priors}
\label{sec:different_priors}

Many regularizers have been proposed to improve the quality of activation maximization images. Here, we compare the benefits of incorporating some of the leading regularizers from previous activation maximization papers.

\subsection{Total variation}
\label{sec:tv}

Optimizing images via gradient descent to maximize the activation of a neuron only, without any regularization, produces unrecognizable images with overly high frequency information~\cite{yosinski2015understanding, nguyen2015deep}. \citet{yosinski2015understanding} proposed incorporating Gaussian blur with a small radius to smooth out the image. While this technique improves recognizability, it causes an overly blurry image because sharp edges are blurred away (Fig.~\ref{fig:comparison}c). Total variation (TV) regularization is a different smoothing technique that combats this issue by minimizing the total variation across adjacent pixels in an image while preserving the sharpness of edges ~\cite{strong2003edge}. In combination with other regularizers such as $\alpha$-norm, this prior is effective in both code inversion and activation maximization tasks~\cite{mahendran2015visualizing}.

While~\citet{mahendran2015visualizing} incorporated the TV norm as a penalty in the objective function (Eq.~\ref{eq:activation_maximization}), here we solve the TV minimization problem separately from the activation maximization problem. Specifically, each of our optimization iterations has two steps: (1) update the image $\mathbf{x}$ in the direction that maximizes the activation of a given neuron; (2) find an image that is closest to $\mathbf{x}$ that has the smallest TV via an iterative split Bregman optimization~\cite{goldstein2009split}.
Split Bregman is an algorithm that is designed to flexibly solve non-differential convex minimization problems by splitting the objective terms, and is especially efficient for TV regularization~\cite{getreuer2012rudin}.
For each iteration of maximizing the neural activation, we perform $100$ iterations of minimizing TV. Empirically, we found this strategy more flexible, which allowed us to discover slightly better results than the method in \citealp{mahendran2015visualizing}.

In extensive preliminary activation maximization experiments (data not shown), we found that fairly good visualizations emerge by applying TV regularization \emph{only}. Our results with TV alone (Fig.~\ref{fig:comparison}g) were not qualitatively improved by adding jitter (Fig.~\ref{fig:comparison}e), and were better than methods that predate \citealp{mahendran2015visualizing} (Fig.~\ref{fig:comparison}b-d).


\subsection{Jitter}
\label{sec:jitter}

An optimization regularization technique called ``jitter'' was first introduced by~\citealp{mordvintsev2015inceptionism} and later used in~\citealp{mahendran2015visualizing}. The method involves: (1) creating a canvas (e.g. of size $272\times272$) that is larger than the DNN input size ($227\times227$) and (2) iteratively optimizing random $227\times227$ regions on the canvas. Optimization with jitter often results in high-resolution, crisp images; however, it does not ameliorate the problems of unnatural coloration and the repetition of image fragments that do not form a coherent, sensible whole. 

Because we believe it represents the best previous activation maximization technique, we reproduce the algorithm of~\citet{mahendran2015visualizing}, which applies both jitter and TV regularizers (Figs.~\ref{fig:mv_only} \&~\ref{fig:comparison}e).  \citet{mahendran2015visualizing} report that TV is the most important prior in their framework. We found that Gaussian blur works just as well if combined with jitter while gradually reducing the blurring effect (i.e. radius) during optimization (Fig.~\ref{fig:comparison}h). This combination works because: (1) as shown in \citet{yosinski2015understanding}, the smoothing effect by Gaussian blur enables optimization to find better local optima that have more global structure; (2) reducing the blur radius over time minimizes the blurring artifact of this prior in the final result; and (3) jitter further enhances the sharpness of images.

\subsection{Center-biased regularization}
\label{sec:center_biased}

Previous activation maximization methods produce images with many unnatural repeated image fragments (Fig.~\ref{fig:comparison}b-h). Though to a lesser degree, multifaceted feature visualization also produces such repetitions (Fig.~\ref{fig:comparison}k). Such repetitions are not found in the vast majority of training set images. For example, a canonical training set image in the ``beacon'' class often shows a single lighthouse (Fig.~\ref{fig:comparison}a); however, Deep Visualization techniques show many more beacons or patches of beacons (Fig.~\ref{fig:comparison}b-h, k). To ameliorate this issue, in this section we introduce a technique called \emph{center-biased regularization}.

\begin{figure}[htb]
	\centering
	\includegraphics[width=0.5\textwidth]{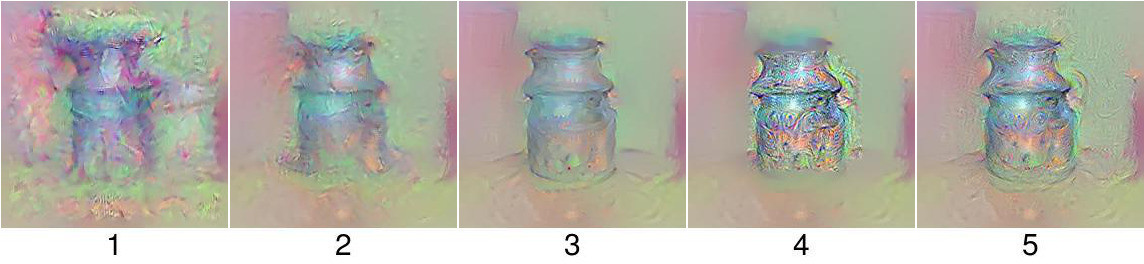}
	\vspace{-0.6cm}
	\caption{Progressive result of optimizing an image to activate the ``milk can'' neuron via \emph{center-biased regularization}. Each image is the result of one out of five optimization phases. This figure is also shown in the main text (Fig.~\ref{fig:series}).
	}
	\label{fig:series_SI}
	\vspace{-0.3cm}
\end{figure}

Our method builds upon the idea of combining TV (Sec.~\ref{sec:tv}) and jitter (Sec.~\ref{sec:jitter}) regularizers, following~\cite{mahendran2015visualizing}, and adds an additional bias towards a restricted ``drawing'' region in the center of the image. In a preliminary experiment (data not shown), we found that optimization with a large smoothing effect (e.g. a high Gaussian blur radius or TV weight) often results in a blurry, but single and centered object in the visualization. Based on this observation, the intuition behind our technique is to first generate a blurry, centered-object image (Fig.~\ref{fig:series_SI}, leftmost image), and then optimize the center pixels more than the edge pixels to produce a final image that is sharp and has a centrally-located image (Fig.~\ref{fig:series_SI}, 4 right images). Multiple examples for different classes are shown in Fig.~\ref{fig:center_bias_only}.

\subsubsection*{Optimization schedule and parameters}

Specifically, our optimization schedule has five phases (Fig.~\ref{fig:series_SI}) that each have a different set of parameters. As described in Sec.~\ref{sec:tv}, each of our optimization iterations has two steps: (1) update the image $\mathbf{x}$ in the direction that maximizes the activation of a given neuron; (2) find a smoothed image $\mathbf{x_s}$ that is closest to $\mathbf{x}$ that has the smallest TV. This $\mathbf{x_s}$ will become the initial image $\mathbf{x}$ in the next optimization iteration. We begin by describing the first three phases, which are the most important. Each runs for 150 iterations.

\begin{figure*}[tb]
	\centering
	\begin{subfigure}{0.49\linewidth}
		\centering
		\includegraphics[width=1.0\linewidth]{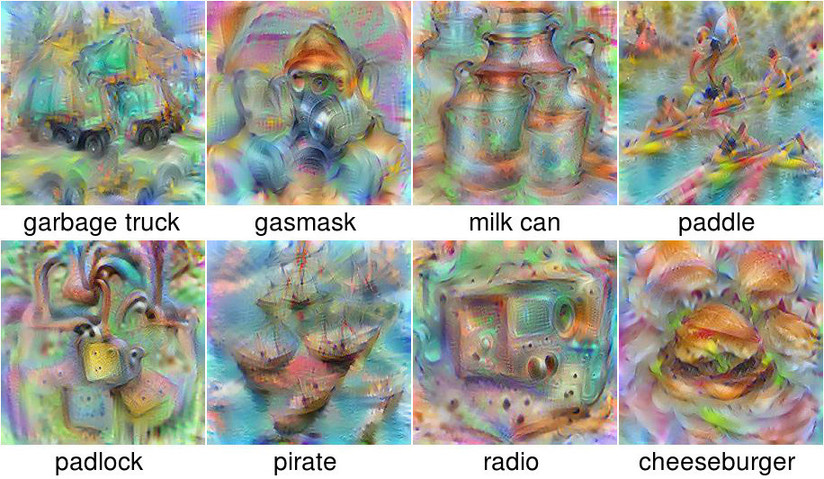}
		\caption{Total variation + Jitter~\cite{mahendran2015visualizing}}
		\vspace{0.2cm}
		\label{fig:mv_only}
	\end{subfigure}	
	\hspace{1mm}
	\begin{subfigure}{0.49\linewidth}
		\centering
		\includegraphics[width=1.0\linewidth]{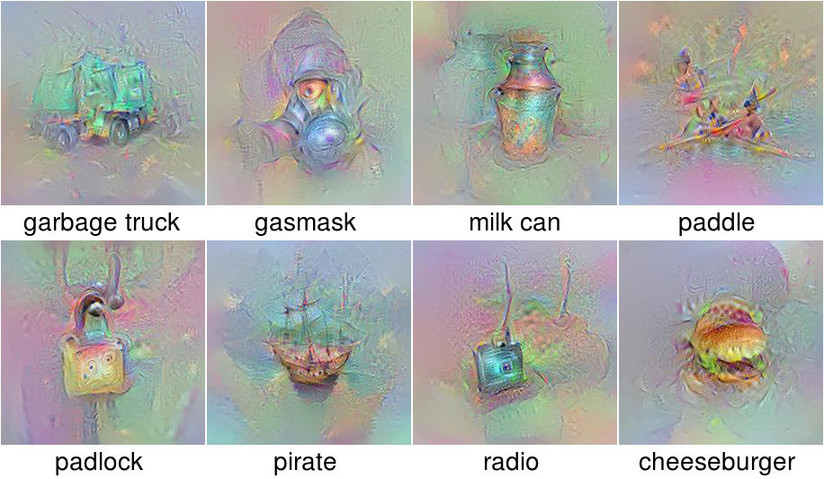}
		\caption{Our \emph{Center-biased regularization} method.}
		\vspace{0.2cm}		
		\label{fig:center_bias_only}
	\end{subfigure}	
	\begin{subfigure}{0.49\linewidth}
		\centering
		\includegraphics[width=1.0\linewidth]{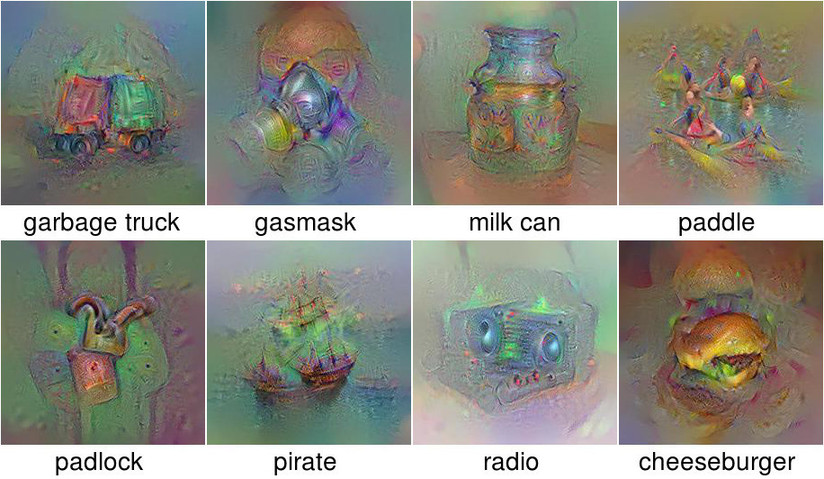}
		\caption{ Center-biased regularization + multifaceted visualization.
		\linebreak\linebreak} 
		\label{fig:mfv_and_center}
	\end{subfigure}
	\hspace{1mm}	
	\begin{subfigure}{0.49\linewidth}
		\centering
		\includegraphics[width=1.0\linewidth]{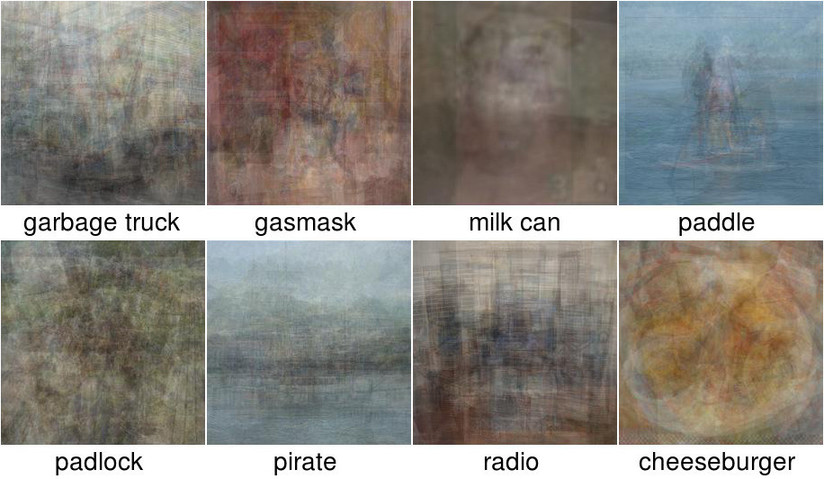}
		\caption{Example mean images that serve as the initialization images for multifaceted feature visualization. Specifically, these were the seeds for the visualizations in (c).}
		\label{fig:mean_images}
	\end{subfigure}
	
	\caption{(a) Previous state of the art activation maximization algorithms produce images with many repeated object fragments that do not form a coherent whole or look natural. (b)
		Center-biased regularization produces images that have fewer repeated fragments, which both represents the general style of images in the training set (one centered object) and make it easier to interpret and understand the feature the neuron in question detects. (c) The realism of the colors of these images are improved when combined with multifaceted visualization.
	}
	\label{fig:center_bias_only_vs_mv}
\end{figure*}

\textbf{Phase 1-3:} First, to generate a blurry image, we start with a low $L_2$ regularization parameter $\lambda=0.001$ when
finding a smoothed image $\mathbf{x_s}$ (step 2). A large
$\lambda$ forces $\mathbf{x_s}$ to be close to $\mathbf{x}$, and results in a sharp image; while a small $\lambda$ allows $\mathbf{x_s}$ to be far from $\mathbf{x}$, and results in a smoothed image. Specifically, for the first 3 phases, $\lambda=\{0.001, 0.08, 2\}$. We also use a lower learning rate for each phase when updating image $\mathbf{x}$ in the activation maximization direction (step 1): $11,6,1$. The intuition is to force the optimization to lock in on the object (e.g. a milk can) that appears in phase 1 (Fig.~\ref{fig:series_SI}, leftmost). In other words, we try to minimize the chance of new duplicated fragments of milk cans to appear toward the end of the optimization as seen in Fig.~\ref{fig:mv_only}.

To bias the main object to appear in the center, for each phase, we increase the canvas size (i.e. upsampling the image $\mathbf{x}$ by $20\%$ at the beginning of each phase) to be: $227\times227$, $272\times272$, and $327\times327$. The regular jittering approach involves sampling and sending a random $227\times227$ patch (anywhere across the canvas) to the DNN for optimization. Here, we restrict such sampling so that the center of a patch is within a small canvas-centered square. By the end of phase 3, the visualization often has a single, centered object (Fig.~\ref{fig:series_SI}, phase 3).

\textbf{Phase 4-5:} The purpose of phase 4 and 5 is to sharpen the centered object without generating new duplicated fragments. We attempt to do this by center-cropping the gradient image (i.e. the gradient backpropagated from the DNN has the form of a $3\times227\times227$ image) down to $3\times127\times127$. In addition, in phase 4, we restrict the optimization to the center of the image only, and optimize for 30 iterations. That sharpens the region in the image center (Fig.~\ref{fig:series_SI}, phase 4). Finally, to balance out the sharpness between the center and edge pixels, in phase 5, we optimize for 10 iterations while allowing jittering to occur anywhere in the image. Thus, the final visualization is often a sharper version of the phase 3 result (Fig.~\ref{fig:series_SI}, phase 5 vs phase 3).

The center-biased regularized images have far fewer duplicated fragments (Fig.~\ref{fig:comparison}i). They thus more closely represent the style of training set images, which feature one centrally located object. However, this technique is not guaranteed to produce a single object only. Instead, it is a way of biasing optimization toward creating objects near the image center. Sec.~\ref{sec:start_mean_images}
shows the experiment combining center-biased regularization and multifaceted visualization (initializing from mean images) to further improve the visualization quality.

\subsection{Initialization with mean images}
\label{sec:start_mean_images}

An issue with previous activation maximization methods is that the images tend to have an unrealistic color distribution (Fig.~\ref{fig:comparison}b-c, e-h). A straightforward approach to ameliorate this problem is to incorporate an $\alpha$-norm regularizer, which encourages the intensity of pixels to stay within a given range~\cite{mahendran2015visualizing,yosinski2015understanding}. While this method is effective in suppressing extreme color values, it does not improve the realism of the overall color distribution of images. \citealp{wei2015understanding} proposed a more advanced data-driven patch prior regularization to enforce the visualizations to match the colors of a set of natural image patches. While this prior substantially improved the colors for code inversion, its results for activation maximization still have several issues (as seen in Fig.~\ref{fig:comparison}d): (1) having duplicated fragments (e.g. duplicated patches of lighthouses in a ``beacon'' image), 
and (2) lacking details, producing unnatural images.

Here, our multifaceted visualization improves the color distribution of images via a different approach: starting optimization from a mean image (Fig.~\ref{fig:mean_images}) computed from real training examples (see Sec.~\ref{sec:mfv_improves_am}). A possible explanation for why this works is that initializing from the mean image puts the search in a low-dimensional manifold that is much closer to that of real images, and thus it is easier to find a realistic looking image around this area. The mean image thus provides a general outline for a type of image based on blurry colors and optimization can fill in the details. 
For example, the center-biased regularization technique was able to produce a single milk can, but did so without a relevant contextual setting (Fig.~\ref{fig:center_bias_only}). However, when this technique is initialized with a mean image (computed from 15 images from the ``milk can'' class) that has a blurry brown surface (Fig.~\ref{fig:mean_images}, milk can), optimization turned that general layout into a complete, coherent picture: a milk can on a table (Fig.~\ref{fig:mfv_and_center}). The colors of the visualizations are also substantially improved when optimization is initialized from mean images (Fig.~\ref{fig:center_bias_only} versus Fig.~\ref{fig:mfv_and_center}).

We observed that multifaceted visualization images often exhibit centered objects (Fig.~\ref{fig:teaser} \&~\ref{fig:facets}), and thus we found no substantial qualitative improvement when adding center-biased regularization to it (Fig.~\ref{fig:mfv_and_center}). That could be because the mean image provides enough of a global layout for the style of the image, which in ImageNet is usually a single, centered object. 

While simple, our technique results in images that have qualitatively more realistic colors than previous methods (Figs.~\ref{fig:fc8_layer} \& ~\ref{fig:comparison}k). This technique is also expected to work with any dataset (e.g. grayscale images, or images of a special topic). 

Note that multifaceted feature visualization does not require access to the training set. If the training set is unavailable, one can simply pass any natural images (or other modes of input such as audio if not reconstructing images) to get a set of images (or other input types) that highly activate a neuron. A similar idea was used in~\citet{wei2015understanding}, who built an external dataset of patches that have similar characteristics to the DNN training set.

\clearpage
\section{What are the hidden units in fully connected layers for?}
\label{sec:explain_fc_units}

We reported in the main text (Sec.~\ref{sec:hidden_neurons}) that neurons in hidden, fully connected layers often seem to be an amalgam of very different, abstract concepts. For example, a reconstruction of one of the facets of a neuron in~\layer{fc6} looks like a combination of a scuba diver and a turtle (Fig.~\ref{fig:all_layers}, leftmost~\layer{fc6} neuron). Here, we document a series of visualization experiments that we performed to further shed light on the inner-workings of hidden, fully connected neurons (in our model, the neurons on \layer{fc6} and \layer{fc7}), such as those visualized in Fig.~\ref{fig:all_layers}. 

\begin{figure*}
	\centering
	\begin{subfigure}{1.0\textwidth}
		\centering
		\includegraphics[width=1.0\linewidth]{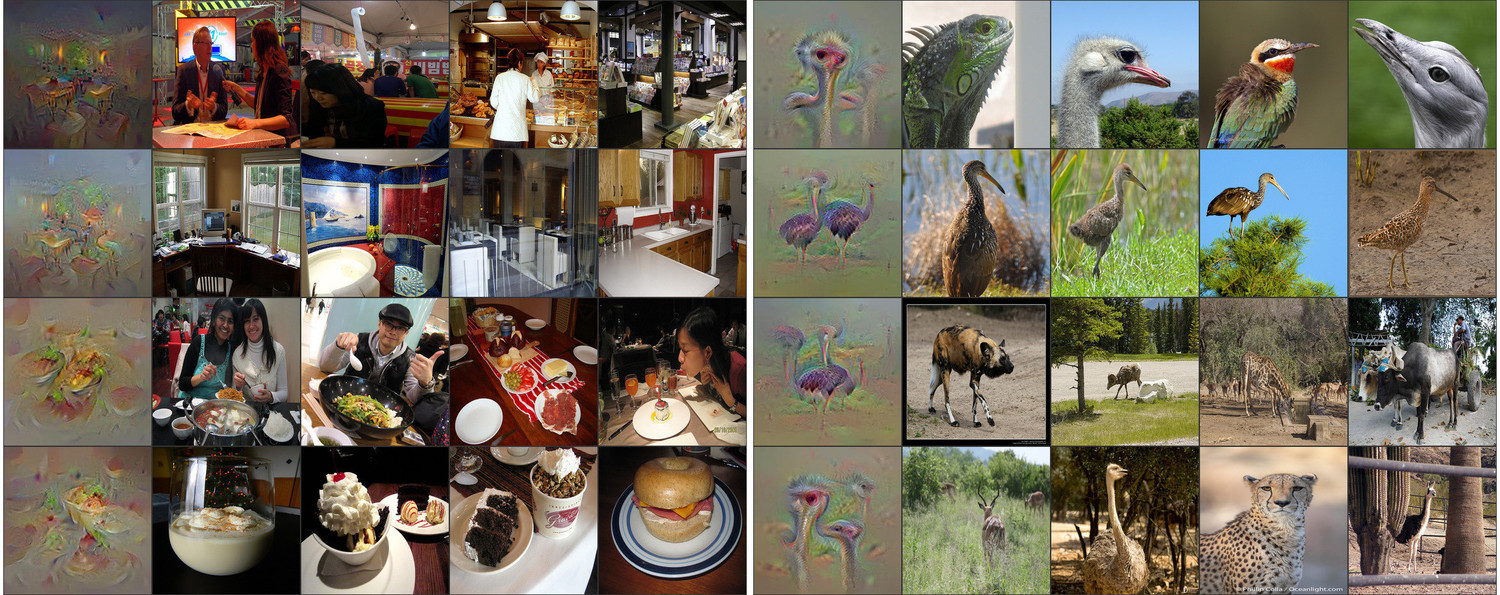}
		\vspace*{-5mm}
		\caption{\layer{fc8} units: ``restaurant'' (left) and ``ostrich'' (right).
		}
		\vspace*{1.5mm}
		\label{fig:fc8}
	\end{subfigure}
	
	\begin{subfigure}{1.0\textwidth}
		\centering
		\includegraphics[width=1.0\linewidth]{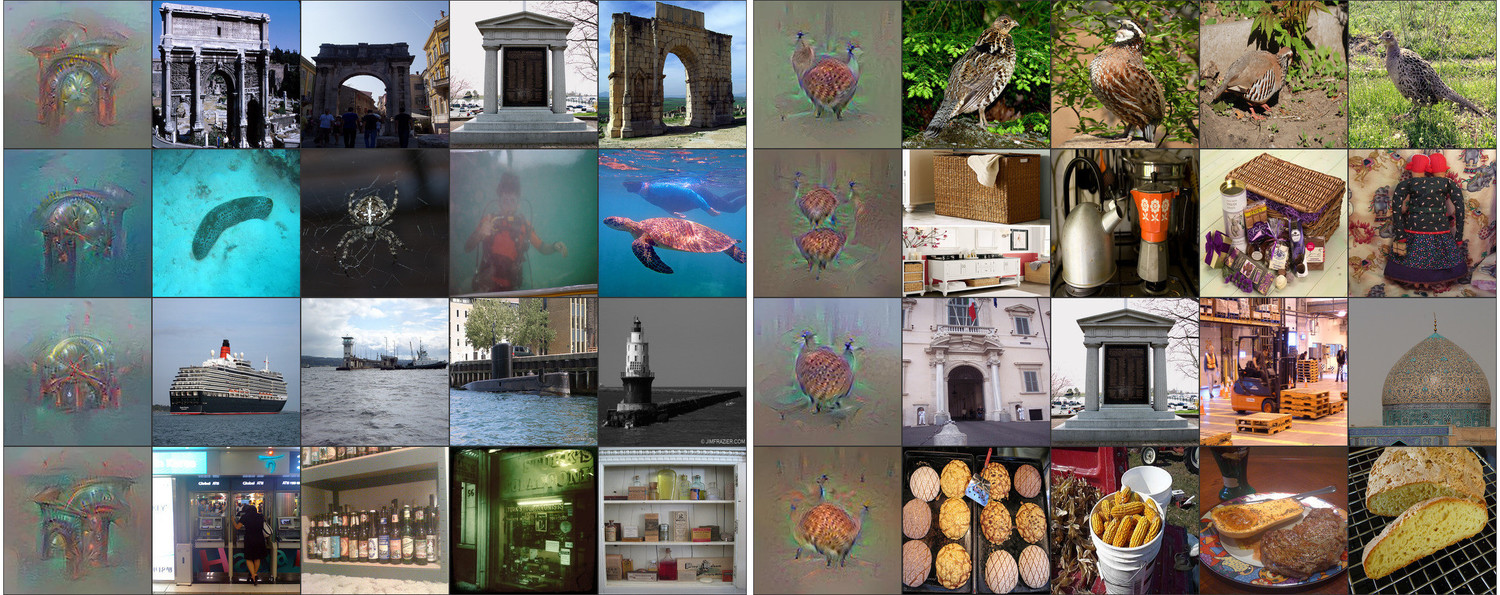}
		\vspace*{-5mm}
		\caption{\layer{fc7} units indexed $159$ (left) and $466$ (right).
		}
		\vspace*{1.5mm}
		\label{fig:fc7}
	\end{subfigure}
	
	\begin{subfigure}{1.0\textwidth}
		\centering
		\includegraphics[width=1.0\linewidth]{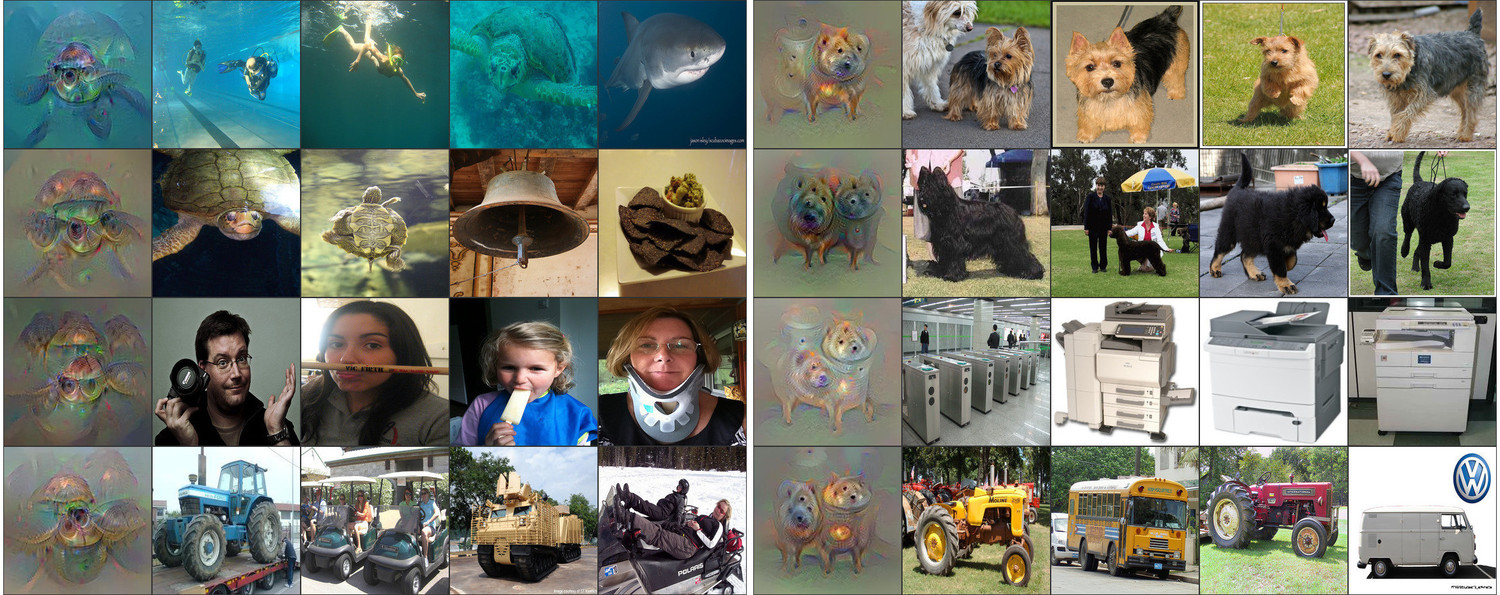}
		\vspace*{-5mm}
		\caption{\layer{fc6} units indexed $22$ (left) and $461$ (right).
		}
		\vspace*{1.5mm}
		\label{fig:fc6}
	\end{subfigure}
	
	\caption{
		To further shed light on the facet reconstructions of~\layer{fc6} and \layer{fc7} neurons discussed in the main text (Fig.~\ref{fig:all_layers}), we investigate two neurons from each of those layers. For each neuron, we show 4 different facets (one per row). For each facet (row), the leftmost image is produced by multifaceted feature visualization; next to it are 4 random images from the 15 validation set images that are used for computing the mean image for that facet. Unlike \layer{fc8} units, the hidden fully connected neurons on \layer{fc6} and \layer{fc7} tend to respond to a variety of different concepts (e.g. from animals to automobiles), and all of their facet reconstructions often converge to similar images.
	}
	\label{fig:intermed_units}
\end{figure*}

The reconstructions for neurons in hidden layers (Fig.~\ref{fig:all_layers}) were produced by running Algorithm~\ref{alg:multifaceted} with $k=10$ clusters on the top 1000 validation set images that most highly activate each hidden neuron.
To understand what feature a neuron fires in response to, we can look at the different types of images that highly activate it (i.e. its different facets). Fig.~\ref{fig:intermed_units} shows, for each cluster of images that highly activate a neuron, 4 random images from that cluster. Specifically, we show 4 random images from the set of 15 images that were averaged to compute the mean image that optimization is initialized with when visualizing that facet. This visualization method is similar to the approach of visualizing the top 9 images that activate a neuron from~\citet{zeiler2014visualizing}, except it is per facet. 

We observe that many individual \layer{fc6} and \layer{fc7} neurons fire for very different types of images. For example, the \layer{fc6} unit that resembles a combination of a scuba diver and a turtle (Fig.~\ref{fig:fc6}, left neuron) does indeed fire for turtles and scuba divers, but also for sharks, bells, human faces and even trucks. Even when multifaceted feature visualization is seeded with a mean image of very different concepts, such as human faces or automobiles, optimization for this neuron consistently converges to very similar images that resemble a turtle with goggles (Fig.~\ref{fig:fc6}, left).

In contrast, and as we would expect given that they are trained to fire in response to one type of image, neurons on the \layer{fc8} layer more clearly represent the same semantic concept. For these neurons, unlike with \layer{fc6} and \layer{fc7} neurons, multifaceted feature visualization produces very unique facet reconstructions for \layer{fc8} neurons that represent different facets of the same semantic concept (Fig.~\ref{fig:fc8_facets}).
For example, the ``restaurant'' class neuron responds to interior views of an entire restaurant in daylight and at night (Fig.~\ref{fig:fc8}, top two rows), and to close-up views of  plates of food (Fig.~\ref{fig:fc8}, bottom two rows). In these cases, and also for convolutional layers,  multifaceted feature reconstructions closely reflect the content of the real images in each facet. In other words, optimization stays near the initial image it is seeded with. 

In some cases, however, even \layer{fc8} neurons respond to an odd assortment of different types of images. 
For example, the ``ostrich'' class neuron fires highly for many non-ostrich images like leopards and lizards (Fig.~\ref{fig:fc8}, right panel).  
Even when viewing the top 9 images that activate a class neuron (ignoring facets), there is usually at least one image from another class, even though there are at least 50 images from that class in the $50,000$-image validation set. Because our MFV algorithm performs clusters on far more than 9 images, some of these clusters will represent an entire group of images that are not of that class (e.g. a ``lizard'' facet for the ostrich neuron). Often, when MFV initializes optimization with the mean image of those non-class images, optimization ``walks away'' from that starting point to produce an image of the class: many examples of this are shown in Fig.~\ref{fig:intermed_units}, such as optimization producing images of ostriches even when seeded with a mean image from a facet of non-ostrich birds or even a facet composed mostly of wild dogs, cows, and giraffes. 

\begin{figure*}[tb]
	\centering
	\begin{subfigure}{0.495\linewidth}
		\centering
		\includegraphics[width=1.0\linewidth]{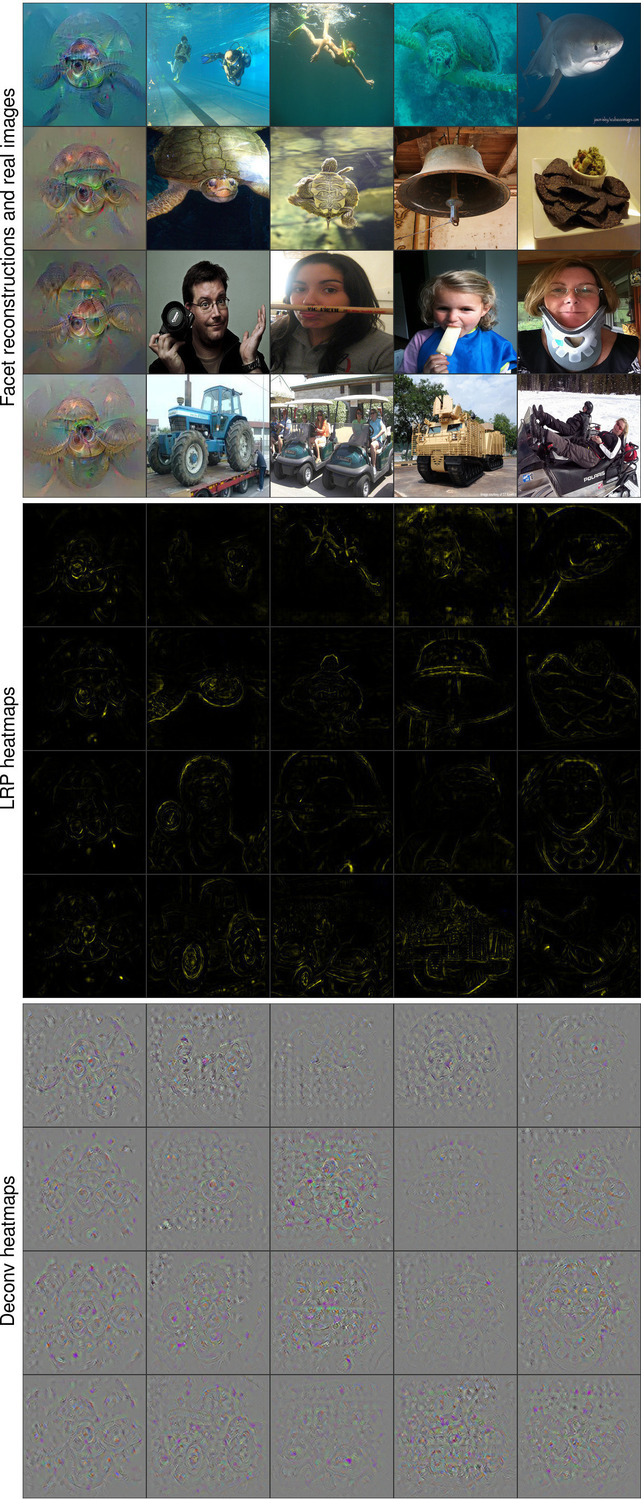}
		\caption{\layer{fc6} unit that resembles ``a scuba diver and a turtle''.}
		\label{fig:lrp_fc6}
	\end{subfigure}	
	\begin{subfigure}{0.495\linewidth}
		\centering
		\includegraphics[width=0.969\linewidth]{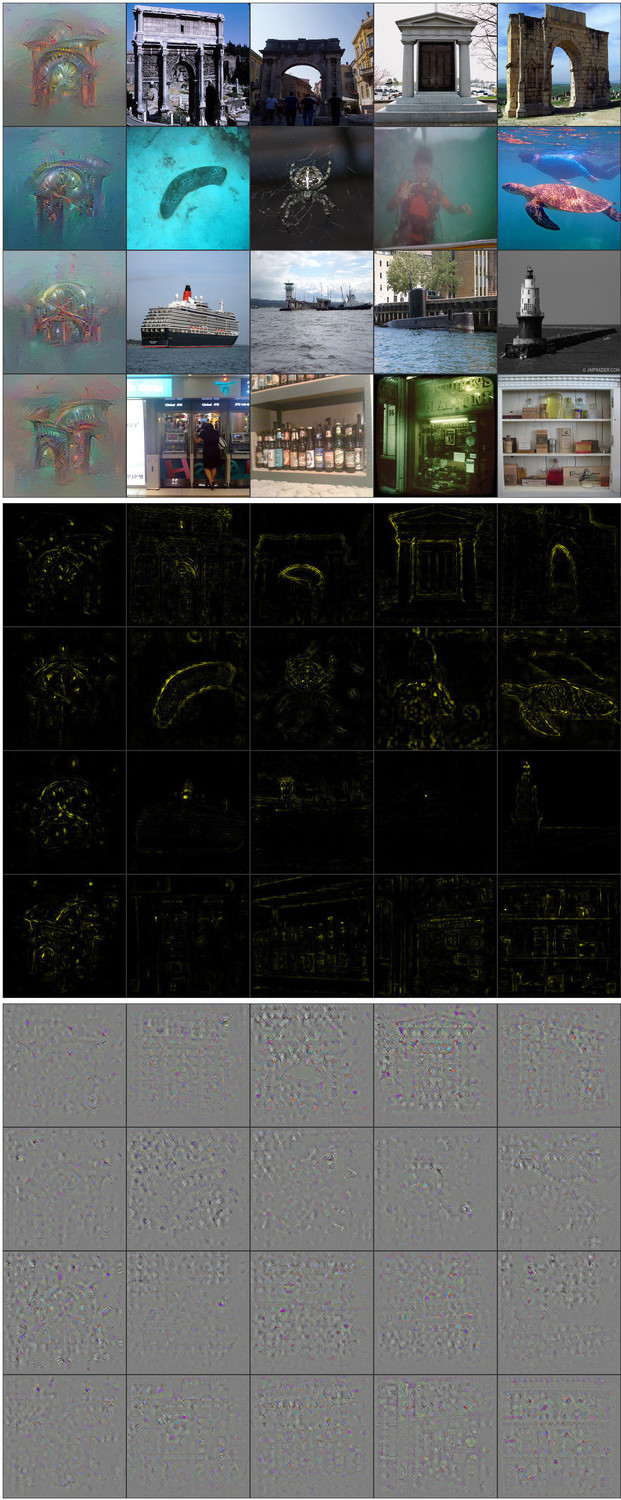}
		\caption{\layer{fc7} unit that resembles an arch.}	
		\label{fig:lrp_fc7}
	\end{subfigure}	

	\caption{
		Visualizing the pixels that are responsible for the activations of an example neuron from \layer{fc6} and another from \layer{fc7} via Layer-wise Relevance Propagation (middle panels)~\cite{bach2015pixel} and deconvolution (bottom panels)~\cite{zeiler2014visualizing}. See the caption of Fig.~\ref{fig:intermed_units} for a description of the top panels.	
		Layer-wise Relevance Propagation heatmaps show that these two neurons indeed fire for a variety of objects, supporting any of the following hypotheses: that the neurons represent an amalgam of different concepts, are part of a distributed representation, or represent a non-obvious abstract concept such as ``curved lines'' or ``curved lines in the vague shape of goggles.'' The Deconv visualizations are quite noisy, making it hard to conclusively identify the most important regions in each image.
	}
	\label{fig:lrp_deconv}
\end{figure*}

While informative, these cluster images could also be misleading because it is not clear whether a given neuron fires for very different types of images (Fig.~\ref{fig:fc6}, photocopiers, trucks, dogs, etc.) or whether that neuron instead fires because there is a common feature present in those different images. We attempted to visualize the pixels in each image that are most responsible for that neuron firing via the deconvolution technique~\cite{zeiler2014visualizing} and via Layer-wise Relevance Propagation (LRP)~\cite{bach2015pixel} (Fig.~\ref{fig:lrp_deconv}). The Deconv results are produced via the DeepVis toolbox~\cite{yosinski2015understanding}, and the LRP results are produced by the LRP toolbox~\cite{bach2015lrp}.

Unfortunately, deconvolution visualizations often contain a lot of noise that makes it hard to identify the most important regions in an image (Fig.~\ref{fig:lrp_deconv}, bottom row). This observation agrees with a recent evaluation of deconv~\cite{samek2015evaluating}. For this reason, we also visualize the images with LRP algorithm. The LRP heatmaps are also difficult to interpret. For example, for the \layer{fc6} ``scuba diver and turtle'' neuron, they show that the outlines of very different shapes cause this neuron to fire, including shapes of underwater things like scuba divers, sharks and turtles (Fig.~\ref{fig:lrp_fc6}, top row in LRP panel), but also a metal bell, human heads, and automobile wheels (Fig.~\ref{fig:lrp_fc6}, LRP panel). It is possible that the neuron cares about the semantics of those disparate object types, or simply that it fires in response to a circular outline pattern that vaguely resembles goggles, which is present in most of these images (albeit at different scales). Another mutually exclusive hypothesis for why the reconstructions are not easily interpretable is that these neurons are part of highly distributed representations. Ultimately, it is unclear exactly what feature this neuron detects. 

Our conclusion remains the same for many other \layer{fc6} and \layer{fc7} units (see also Fig.~\ref{fig:lrp_fc7}): while it is unclear exactly what they represent, these neurons do seem to represent an amalgam of different types of images, and optimization often produces the same hybrid visualization no matter where it is started from, such as the yellow dog+bathtub on \layer{fc6} (Fig.~\ref{fig:fc6}). Other times there is a dominant visualization that also appears no matter where optimization starts, but that is more semantically homogeneous (instead of being a hybrid of very different concepts), such as an arch or quail-like bird (Fig.~\ref{fig:fc7}).

Overall, the evidence in this section reveals how little we know about the precise function of hidden neurons in fully connected layers, and future research is required to make progress on this interesting question.

\end{document}